%% file: main.tex
\documentclass[nohyperref,hyphens]{article}

\usepackage{microtype}
\usepackage{graphicx}
\usepackage{subfigure}
\usepackage{booktabs} 

\usepackage{hyperref}

\usepackage[accepted]{icml2022}

\usepackage{amsmath}
\usepackage{amssymb}
\usepackage{mathtools}
\usepackage{amsthm}

\usepackage[capitalize,noabbrev]{cleveref}

\theoremstyle{plain}
\newtheorem{theorem}{Theorem}[section]
\newtheorem{proposition}[theorem]{Proposition}
\newtheorem{lemma}[theorem]{Lemma}
\newtheorem{corollary}[theorem]{Corollary}
\theoremstyle{definition}

\newtheorem{assumption}[theorem]{Assumption}
\theoremstyle{remark}

\hypersetup{colorlinks=true,breaklinks=true}

\usepackage{nicefrac}
\usepackage{enumitem}
\setitemize{leftmargin=1em,itemsep=-1pt, topsep=0pt, parsep=0em,label=\raisebox{0.25ex}{\tiny$\bullet$}}
\setlist[enumerate]{leftmargin=*, label= {\arabic*.}, itemsep=0em, topsep=0pt}

\input{math_commands}
\allowdisplaybreaks

\icmltitlerunning{Understanding Instance-Level Impact of Fairness Constraints}

\begin{document}

\twocolumn[
\icmltitle{Understanding Instance-Level Impact of Fairness Constraints}



\icmlsetsymbol{equal}{*}

\begin{icmlauthorlist}
\icmlauthor{Jialu Wang}{ucsc}
\icmlauthor{Xin Eric Wang}{ucsc}
\icmlauthor{Yang Liu}{ucsc}
\end{icmlauthorlist}

\icmlaffiliation{ucsc}{Department of Computer Science and Engineering, University of California, Santa Cruz, CA, USA. Email: Jialu Wang \textless{}faldict@ucsc.edu\textgreater{}, Xin Eric Wang \textless{}xwang366@ucsc.edu\textgreater{}, Yang Liu \textless{}yangliu@ucsc.edu\textgreater{}
}

\icmlcorrespondingauthor{Yang Liu}{yangliu@ucsc.edu}

\icmlkeywords{Fairness, Influence Function}

\vskip 0.3in
]



\printAffiliationsAndNotice{}  

\begin{abstract}
    A variety of fairness constraints have been proposed in the literature to mitigate group-level statistical bias. Their impacts have been largely evaluated for different groups of populations corresponding to a set of sensitive attributes, such as race or gender. Nonetheless, the community has not observed sufficient explorations for how imposing fairness constraints fare at an instance level. Building on the concept of influence function, a measure that characterizes the impact of a training example on the target model and its predictive performance, this work studies the influence of training examples when fairness constraints are imposed. We find out that under certain assumptions, the influence function with respect to fairness constraints can be decomposed into a kernelized combination of training examples. One promising application of the proposed fairness influence function is to identify suspicious training examples that may cause model discrimination by ranking their influence scores. We demonstrate with extensive experiments that training on a subset of weighty data examples leads to lower fairness violations with a trade-off of accuracy.
\end{abstract}

\input{Section/Introduction}
\input{Section/Preliminary}

\input{Section/InfluenceFunction}
\input{Section/GeneralizationBound}
\input{Section/Experiment}
\input{Section/Discussion}

\section*{Acknowledgement}
This work is partially supported by the National Science Foundation (NSF) under grants IIS-2007951, IIS-2143895, IIS-2040800 (FAI program in collaboration with Amazon), and CCF-2023495. This work is also partially supported by UC Santa Cruz Applied Artificial Intelligence Initiative (AAII). The authors would like to thank Tianyi Luo for providing data pre-processing scripts on Jigsaw Toxicity data source and the anonymous reviewers for their constructive feedback.

\bibliography{reference}
\bibliographystyle{icml2022}

\newpage
\appendix
\onecolumn

\input{Section/Appendix}


\end{document}

%% file: math_commands.tex
\usepackage{amsmath,amsfonts,bm}
\usepackage{physics}

\def\eqref#1{equation~\ref{#1}}

\def\1{\bm{1}}

\DeclareMathAlphabet{\mathsfit}{\encodingdefault}{\sfdefault}{m}{sl}
\SetMathAlphabet{\mathsfit}{bold}{\encodingdefault}{\sfdefault}{bx}{n}

\def\gF{{\mathcal{F}}}

\def\gX{{\mathcal{X}}}
\def\gY{{\mathcal{Y}}}
\def\gZ{{\mathcal{Z}}}

\newcommand{\E}{\mathbb{E}}

\newcommand{\R}{\mathbb{R}}

\def\Expectation{\mathbb{E}}
\def\parameter{{\boldsymbol\theta}}

\def\data{\mathcal{D}}
\def\dataset{D}

\def\err{\text{err}}
\def\constraint{\phi}
\def\infl{\text{infl}}
\newcommand{\independent}{\perp \!\!\! \perp}

\newcommand\at[2]{\left.#1\right|_{#2}}

\def\SuchThat{s.t.}
\def\Covariance{\mathtt{Cov}}

\DeclareMathOperator{\TPR}{TPR}
\DeclareMathOperator{\FPR}{FPR}


%% file: Section/Introduction.tex
\section{Introduction}
Machine learning models have been deployed in a variety of real-world decision-making systems, including hiring \cite{Ajunwa2016HiringBA,Bogen2018HelpWA,10.1145/3351095.3372828}, loan application \cite{Siddiqi2005CreditRS,10.1145/3287560.3287566}, medical treatments \cite{Pfohl2021AnEC,zhou2021radfusion}, recidivism assessment \cite{angwin2016machinebias,Chouldechova_2016} and more. Nonetheless, a number of studies have reported unfair treatments from a machine learning model \cite{Mayson2018BiasIB,2016COMPASRS,pmlr-v81-buolamwini18a,10.5555/3157382.3157584,10.1145/3287560.3287596}. The fair machine learning community has responded with solutions \cite{corbett2017algorithmic,Feldman2015,chuang2021fair,Taskesen2020ADR}, with the core idea being to impose fairness constraints on either the group level \cite{hardt2016equalodds,fairlearn,DBLP:journals/corr/abs-1802-06309,pmlr-v65-woodworth17a} or the individual level \cite{dwork2012fairness,NIPS2017_a486cd07,NEURIPS2019_e94550c9}.

Despite the successes of the algorithmic treatments, the question of \emph{why} a particular ``fair" training process leads to a more fair model remains less addressed. The explanation for the above \emph{why} question is essential in improving user trustworthiness in the models and often regulated by legal requirements \cite{ModelCivil}. 
There has been a recent surge of interest in explaining algorithmic fairness. Much of the work chose to quantify the importance of the input feature variables used to make fair decisions \cite{NIPS2017_8a20a862,pmlr-v119-sundararajan20b,Mase2021CohortSV}. This line of research makes explanations on the population level, as the importance measures are quantified statistically over the entire subset of instances. 

Nevertheless, the impact of fairness constraints on individual instances is rarely discussed. The central inquiry of this paper is how each individual training instance influences the model decisions when a fairness constraint is imposed.
Demystifying and characterizing the influence of individual instances subject to fairness constraints is important and opens up the possibility of auditing a machine learning model at the instance level. Among other potentials, we believe that such understanding might help with developing preprocessing solutions to mitigate bias by emphasizing more on instances that have a high influence on fairness.

To this end, we borrow the idea from recent literature on \emph{influence function} \cite{pmlr-v70-sundararajan17a}, which has largely focused on approximating the effect of training examples in prediction accuracy rather than fairness constraints. Concretely, an influence function characterizes the change of model predictions compared to the counterfactual that one training example is removed. We instantiate the change, due to the penalty of disparity, on prominent fairness criteria that have been widely applied in the community. We illustrate that the influence scores can be potentially applied to mitigate the unfairness by pruning less influential examples on a synthetic setting. We implement this idea on different domains including tabular data, images and natural language.

\subsection{Related Work}
Our work is mostly relevant to the large body of results on algorithmic fairness. Diverse equity considerations have been formulated by regulating the similar treatments between similar individuals \cite{dwork2012fairness}, comparing the outcome of an individual with a hypothetical counterpart who owns another sensitive attribute \cite{counterfactual_fairness}, enforcing statistical regularities of prediction outcomes across demographic groups \cite{Feldman2015,Chouldechova_2016,hardt2016equalodds}, or contrasting the performance of subpopulations under semi-supervision \cite{zhu2022the} or zero-shot transfer \cite{wang-etal-2022-assessing}. This work mainly focuses on group-based fairness notions, including demographic disparity \cite{Chouldechova_2016} and equality of opportunity \cite{hardt2016equalodds}.

The theoretical analysis posed in this work is grounding on practical algorithms for mitigating group fairness violations. While we observe that the approaches for developing a fair model broadly include reweighing or distorting the training instances \cite{KamiranFaisal2012DataPT,Feldman2015,NIPS2017_6988,liu2021can} and post-process the models to correct for the discrimination \cite{hardt2016equalodds,petersen2021postprocessing}, we will focus on the solutions that incorporate the fairness constraints in the learning process \cite{Cotter2019OptimizationWN,Zafar2017FairnessCM,pmlr-v65-woodworth17a,fairlearn,Song2019LearningCF,Kamishima2011FairnessawareLT,10.1145/3442188.3445915}.

This work is closely related to the line of research in influence function and memorization \cite{Feldman2020DoesLR,NEURIPS2020_1e14bfe2}. Influence functions \cite{Cook1980CharacterizationsOA} can be used to measure the effect of removing an individual training instance on the model predictions at deployment. In the literature, prior works formulate the influence of training examples on either the model predictions \cite{pmlr-v70-sundararajan17a,pmlr-v70-koh17a} or the loss \cite{tracin}. In our paper, we are interested in the influence subject to fairness constraints on both the model predictions and performance, with a focus on the former. Prior works \cite{pmlr-v70-koh17a,tracin} have shown a first-order approximation of influence function can be useful to interpret the important training examples and identify outliers in the dataset. There is also an increasing applications of influence function on tasks other than interpretability \cite{DBLP:conf/icml/BasuYF20,basu2021influence}. In NLP, influence function have been used to diagnose stereotypical biases in word embeddings \cite{pmlr-v97-brunet19a}. In security and robustness, attackers can exploit influence function \cite{DBLP:journals/corr/abs-1811-00741} to inject stronger poisoned data points. In semi-supervised learning, influence function can be deployed to identify the examples with corrupted labels \cite{zhu2022detect}. A recent work demonstrates the efficacy of training overparametrized networks on the dataset where a large fraction of less important examples are discarded by computing self-influence \cite{paul2021deep}. We mirror their idea of utilizing influence function to prune data examples and illustrate with experiments that the model trained on a subset of training data can have a lower fairness violations.

Our desire to explore the effect of removing an individual from the training set is also parallel to a recent work on leave-one-out unfairness \cite{black2021loo}. Our work primarily differs in two aspects. Firstly, leave-one-out unfairness focuses on formalizing the stability of models with the inclusion or removal of a single training point, while our work aims to measure the influence of imposing a certain fairness constraint on an individual instance. Secondly, we explicitly derive the close-form expressions for the changes in either model outputs or prediction loss to a target example.

\subsection{Our Contributions}
The primary contribution of this work is to provide a feasible framework to interpret the impact of group fairness constraints on individual examples. Specifically, we develop a framework for estimating the influence function with first-order approximation (see Section \ref{sec:influence-function}). We postulate the decomposability of fairness constraints, and pose a general influence function as the product of a kernel term, namely neural tangent kernel, and a gradient term related to the specific constraints (see Lemma \ref{lem:constrained-influence}). We instantiate the concrete influence function on a variety of exemplary fairness constraints (see Section \ref{sec:exemplary-fairness-constraint}). As a direct application, we demonstrate that the influence scores of fairness constraints can lend itself to prune the less influential data examples and mitigate the violation (see Section \ref{sec:evaluation}). We defer all subsequently omitted proofs to Appendix \ref{appendix:omitted-proof}. We publish the source code at \url{https://github.com/UCSC-REAL/FairInfl}.

%% file: Section/Preliminary.tex
\section{Preliminary}
We will consider the problem of predicting a target binary label $y$ based on its corresponding feature vector $x$ under fairness constraints with respect to sensitive attributes $z$. We assume that the data points $(x, y, z)$ are drawn from an unknown underlying distribution $\data$ over $\gX \times \gY \times \gZ$. $\gX \in \R^d$ is $d$-dimensional instance space, $\gY \in \{-1, +1\}$ is the label space, and $\gZ \in \{0, 1, \ldots, m-1\}$ is the (sensitive) attribute space. Here we assume that sensitive attribute is a categorical variable regarding $m$ sensitive groups. The goal of fair classification is to find a classifier $f : \gX \to \R$ with the property that it minimizes expected true loss $\err(f)$ while mitigating a certain measure of fairness violation $\psi(f)$.
\begin{table*}[!htb]
    \centering
    \caption{Examples of fairness measures.}
    \label{tab:example-fairness-constraint}
    \begin{tabular}{l c l}
        \toprule
            Fairness Criteria & & Measure $\psi(f)$ \\
        \midrule 
            Demographic Parity & & $\sum_{g \in \gZ} |\Pr(f(x) = +1 \mid z = g) - \Pr(f(x) = +1)|$ \\
            Equal True Positive Rate & & $\sum_{a \in \gZ} |\Pr(f(x) = +1 \mid z = a, y = +1) - \Pr(f(x) = +1 \mid y = +1)|$ \\
            Equal False Positive Rate & & $\sum_{a \in \gZ} |\Pr(f(x) = +1 \mid z = a, y = -1) - \Pr(f(x) = +1 \mid y = -1)|$ \\
            Equal Odds & & $\sum_{a \in \gZ}\sum_{b \in \gY} |\Pr(f(x) = +1 \mid z = a, y = b) - \Pr(f(x) = +1 \mid y = b)|$\\
        \bottomrule
    \end{tabular}
\end{table*}
We assume that the model $f$ is parameterized by a vector $\parameter = [\theta_1, \theta_2, ..., \theta_p]$ of size $p$. Thereby $\err(f) = \E_{(x, y) \sim \data} [\ell (f(x; \parameter), y)]$, where the expectation is respect to the true underlying distribution $\data$ and $\ell(\cdot, \cdot)$ is the loss function. We show exemplary fairness metrics $\psi(\cdot)$ in Table \ref{tab:example-fairness-constraint}, including demographic parity~\cite{Chouldechova_2016,jiang2022generalized}, equality of opportunity~\cite{hardt2016equalodds}, among many others.
Without loss of generality, $f(x)$ induces the prediction rule $2 \cdot \1 [f(x) \geq 0] - 1$, where $\1[\cdot]$ is the indicator function. Denote by $\gF$ the family of classifiers, we can express the objective of the learning problem as
\begin{equation}\label{eq:Risk}
    \min_{f \in \gF}~\err(f), ~~\SuchThat~\psi(f) \leq \mu,
\end{equation}
where $\mu$ is a tolerance parameter for fairness violations. Let $\dataset = \{(x_i, y_i, z_i)\}_{i=1}^n$ denote $n$ data examples sampled from true data distribution $\data$. In this case, the empirical loss is $\widehat{\err}(f) = \frac{1}{n} \sum_{(x_i, y_i, z_i) \in D} \ell(f(x_i), y_i)$. Due to the fact that $\psi(f)$ is non-convex and non-differentiable in general, practically we will use a surrogate $\constraint(f)$ to approximate it. We will defer the examples of $\constraint(f)$ until Section \ref{sec:exemplary-fairness-constraint}. Let $\hat{\constraint}(\cdot)$ denote the empirical version of $\constraint(\cdot)$, then the \emph{Empirical Risk Minimization} (ERM) problem is defined as
\begin{equation}\label{eq:ERM}
    \min_{f \in \gF}~\widehat{\err}(f), ~~\SuchThat~\hat{\constraint}(f) \leq \mu.
\end{equation}
This work aims to discuss the influence of a certain training example $(x_i, y_i, z_i)$ on a target example $(x_j, y_j, z_j)$, when fairness constraints are imposed to the classifier $f$. Let $f^{D}$ represent the model $f$ trained over the whole dataset $D$ and $f^{D/\{i\}}$ represent the counterfactual model $f$ trained over the dataset $D$ by excluding the training example $(x_i, y_i, z_i)$. The influence function with respect to the output of classifier $f$ is defined as
\begin{equation}\label{defi:influence-function}
    \infl_f (D, i, j) \coloneqq f^{D/\{i\}} (x_j) - f^{D}(x_j)
\end{equation}
Note that $j$ may be either a training point $j \in D$ or a test point outside $D$.

%% file: Section/InfluenceFunction.tex
\section{Influence Function}\label{sec:influence-function}
The definition of influence function stems from a hypothetical problem: how will the model prediction change compared to the counterfactual that a training example is removed? Prior work on influence function has largely considered the standard classification problem where a certain loss function is minimized. In this section, we will firstly go over the formulation of influence function in this unconstrained setting. We follow the main idea used by \cite{pmlr-v70-koh17a} to approximate influence function by first-order Taylor series expansion around the parameter $\parameter$ of model $f$. Then we extend the approximated influence function into the constrained setting where the learner is punished by fairness violations.

\subsection{Influence Function in Unconstrained Learning}\label{sec:unconstrained-influence}
We start by considering the unconstrained classification setting when parity constraints are not imposed in the learning objective. Recall that the standard Empirical Risk Minimization (ERM) problem is
\begin{align}
    \min_\parameter \frac{1}{n}\sum_{i=1}^n\ell(f(x_i; \parameter), y_i) 
\end{align}
where $\parameter$ is the parameters of model $f$. 
We assume that $\parameter$  evolves through the following gradient flow along time $t$:
\begin{align}\label{eq:ode}
\frac{\partial \parameter}{\partial t} = - \frac{1}{n} \nabla \ell(f(x_i; \parameter), y_i)
\end{align}
Let $\parameter_0$ denote the final parameter of classifier $f$ trained on the whole set $D$. To track the influence of an observed instance $i$, we hypothesis the update of parameter $\parameter$ with respect to instance $i$ is recovered by one counterfactual step of gradient descent with a weight of $\nicefrac{-1}{n}$ and a learning rate of $\eta$. This process can also be regarded as inverting the gradient flow of Equation \ref{eq:ode} with a small time step $\Delta t = \eta$. Next, to compute the output of model $f$ on the target example $j$, we may Taylor expand $f$ around $\parameter_0$
\begin{align}
    & f(x_j; \parameter) - f(x_j; \parameter_0) \nonumber \\
    \approx & \frac{\partial f(x_j; \parameter_0)}{\partial \parameter} (\parameter - \parameter_0) \nonumber \tag{by Taylor series expansion} \\
    = & \frac{\partial f(x_j;\parameter_0)}{\partial \parameter} \left(-\eta \at{\frac{\partial \parameter}{\partial t}}{\parameter = \parameter_0}\right) \nonumber \tag{by inverting gradient flow} \\ 
    = & \frac{\eta}{n} \frac{\partial f(x_j;\parameter_0)}{\partial \parameter} \nabla \ell (f(x_i;\parameter_0), y_i) \nonumber \tag{by substituting Equation \ref{eq:ode}}\\
    = & \frac{\eta}{n} \frac{\partial f(x_j;\parameter_0)}{\partial \parameter} \frac{\partial \ell(f(x_i;\parameter_0),y_i)}{\partial f} \frac{\partial f(x_i; \parameter_0)}{\partial \parameter} \nonumber\tag{by chain rule} \\
    = & \frac{\eta}{n} \frac{\partial f(x_j;\parameter_0)}{\partial \parameter}\frac{\partial f(x_i; \parameter_0)}{\partial \parameter} \at{\frac{\partial\ell(w, y_i)}{\partial w}}{w = f(x_i; \parameter_0)} \label{eq:derive-influence-function}
\end{align}
In the language of kernel methods, the product of $\pdv*{f(x_i; \parameter)}{\parameter}$ and $\pdv*{f(x_j; \parameter)}{\parameter}$ is named Neural Tangent Kernel (NTK) \cite{NEURIPS2018_5a4be1fa}
\begin{equation}\label{eq:ntk}
\begin{aligned}
    \Theta(x_i, x_j; \parameter) & = \frac{\partial f(x_j;\parameter)}{\partial \parameter}\frac{\partial f(x_i; \parameter)}{\partial \parameter} \\ 
    & = \sum_{p} \frac{\partial f(x_j;\parameter)}{\partial \theta_p}\frac{\partial f(x_i; \parameter)}{\partial \theta_p} 
\end{aligned}
\end{equation}
NTK describes the evolution of deep neural networks during the learning dynamics. Substituting the NTK in Equation \ref{eq:ntk} into Equation \ref{eq:derive-influence-function} and combining Equation \ref{defi:influence-function}, we obtain the following close-form statement: 
\begin{lemma}[]\label{lem:unconstrained-influence}
In unconstrained learning, the influence function of training example $i$ subject to the prediction of $f$ on the target example $j$ is
\begin{equation}\label{eq:unconstrained-influence}
    \infl_f(D, i, j) \approx \frac{\eta}{n} \Theta(x_i, x_j; \parameter_0) \at{\frac{\partial\ell(w, y_i)}{\partial w}}{w = f(x_i; \parameter_0)}
\end{equation}
\end{lemma}
Equation \ref{eq:unconstrained-influence} mimics the first-order approximation in \cite{tracin} with a focus on tracking the change on model output instead of the change on loss.
\subsection{Influence Function in Constrained Learning}
In classification problems, the outcome of an algorithm may be skewed towards certain protected groups, such as gender and ethnicity. While the definitions of fairness are controversial, researchers commonly impose the parity constraints like demographic parity \cite{Chouldechova_2016} and equal opportunity \cite{hardt2016equalodds} for fairness-aware learning. A large number of approaches have been well studied to mitigate the disparity, which in general can be categorized into pre-processing, in-processing, and post-processing algorithms. Pre-processing algorithms \cite{KamiranFaisal2012DataPT,Feldman2015,NIPS2017_6988} usually reweigh the training instances, resulting in the influence scores will also be scaled by a instance-dependent weight factor. Post-Processing algorithms \cite{hardt2016equalodds} will not alter the learning objective, thus the influence function of training examples stays unchanged.

In this work, we primarily focus on the influence function in the in-processing treatment frameworks \cite{Cotter2019OptimizationWN,Zafar2017FairnessCM,pmlr-v65-woodworth17a,fairlearn,pmlr-v84-narasimhan18a,Song2019LearningCF,Kamishima2011FairnessawareLT}. In such fashion, the fair classification problem are generally formulated as a constrained optimization problem as Equation \ref{eq:Risk}. The common solution is to impose the penalty of fairness violations $\psi(f)$ as a regularization term. The constrained risk minimization problem thus becomes
\begin{align}
    \min_{f \in \gF}~\err(f) + \lambda \psi(f) 
\end{align}
where in above $\lambda$ is a regularizer that controls the trade-off between fairness and accuracy. Note $\lambda$ is not necessary static, e.g., in some game-theoretic approaches \cite{fairlearn,pmlr-v84-narasimhan18a,Cotter2019OptimizationWN,Cotter2019TwoPlayerGF}, the value of $\lambda$ will be dynamically chosen. We notice that while the empirical $\psi(f)$ is often involving the rates related to indicator function, it might be infeasible to solve the constrained ERM problem. For instance, demographic parity, as mentioned in Table \ref{tab:example-fairness-constraint}, requires that different protected groups have an equal acceptance rate. The acceptance rate for group $a \in \gZ$ is given by
\[
    \quad \Pr (f(x) \geq 0 \mid z = a) = \frac{\sum_i \1[f(x_i) \geq 0, z_i = a]}{\sum_i \1[z_i = a]}
\]
Since non-differentiable indicator function cannot be directly optimized by gradient-based algorithms, researchers often substitute the direct fairness measure $\psi(f)$ by a differentiable surrogate $\phi(f)$. In consequence, the constrained ERM problem is
\begin{align}
    \min_\parameter \frac{1}{n}\sum_{i=1}^n\ell(f(x_i; \parameter), y_i) + \lambda \hat{\constraint}(f)
\end{align}
We make the following decomposability assumption:
\begin{assumption}[Decomposability]\label{asm:decomposable}
    The empirical surrogate of fairness measure $\hat{\constraint}(f)$ can be decomposed into
    \[
        \hat{\constraint}(f) = \frac{1}{n}\sum_{i=1}^n \hat{\constraint}(f, i),
    \]
    where in above each $\hat{\constraint}(f, i)$ is only related to the instance $i$ and independent of other instances $j \neq i$.
\end{assumption}
Assumption \ref{asm:decomposable} guarantees that the influence of one training example $i$ will not be entangled with the influence of another training example $j$. Following an analogous derivation to Equation \ref{eq:derive-influence-function}, we obtain the kernelized influence function
\begin{lemma}\label{lem:constrained-influence}
    When the empirical fairness measure $\hat{\constraint}(\cdot)$ satisfies the decomposability assumption, the influence function of training example $i$ with respect to the prediction of $f$ on the target $j$ can be expressed as
    \begin{equation}\label{eq:constrained-influence}
        \begin{aligned}
        \infl_f(D, i, j) & \approx \underbrace{\frac{\eta}{n} \Theta(x_i, x_j; \parameter_0) \at{\frac{\partial\ell(w, y_i)}{\partial w}}{w = f(x_i; \parameter_0)}}_{\text{influence of loss}} \\
        & + \underbrace{\lambda \frac{\eta}{n} \Theta(x_i, x_j; \parameter_0) \at{\frac{\partial\hat{\constraint}(f, i)}{\partial f}}{f(x_i; \parameter_0)}}_{\text{influence of fairness constraint}}
        \end{aligned}
    \end{equation}
\end{lemma}
Lemma \ref{lem:constrained-influence} presents that the general expression of influence function can be decoupled by the influence subject to accuracy (the first term) and that subject to parity constraint (the second term). 

The above result reveals the influence function in terms of the change of model outputs. We may also track the change of loss evaluated on a target example. We note prior work \cite{tracin} has proposed effective solutions and will defer more discussions to Appendix \ref{appendix:influence-on-loss}.

\section{Influence of Exemplary Fairness Constraints}\label{sec:exemplary-fairness-constraint}
In this section, we will take a closer look at the specific influence functions for several commonly used surrogate constraints. Since the influence induced by loss is independent of the expressions for fairness constraint, we will ignore the first term in Equation \ref{eq:constrained-influence} and focus on the second term throughout this section. We define the pairwise influence score subject to fairness constraint as
\begin{equation}\label{eq:influence-score}
    S(i, j) \coloneqq \lambda \frac{\eta}{n} \Theta(x_i, x_j; \parameter_0) \at{\frac{\partial\hat{\constraint}(f, i)}{\partial f}}{f(x_i; \parameter_0)}
\end{equation}
In what follows, we will instantiate $S(i, j)$ on three regularized fairness constraints.

\subsection{Relaxed Constraint}
Throughout this part, we assume that the sensitive attribute is binary, i.e., $\gZ \in \{0, 1\}$. The technique of relaxing fairness constraints was introduced in \cite{DBLP:journals/corr/abs-1802-06309}. We will analyse the influence of relaxed constraints, including demographic parity and equality of opportunity as below. 
\paragraph{Demographic Parity.} \citet{DBLP:journals/corr/abs-1802-06309} propose to replace the demographic parity metric
\begin{equation}
\begin{multlined}
    \psi(f) = |\Pr(f(x; \parameter) \geq 0 \mid z = 1) \\ - \Pr(f(x; \parameter) \geq 0 \mid z=0)|
\end{multlined}
\end{equation}
by a relaxed measure
\begin{equation}\label{eq:relax-dp}
    \phi(f) = |\Expectation [f(x; \parameter) \cdot \1[z = 1]] - \Expectation[f(x; \parameter)\cdot \1[z=0]]|
\end{equation}
Without loss of generality, we assume that the group $z=1$ is more favorable than the group $z=0$ such that $\Expectation [f(x; \parameter)\1[z = 1]] \geq \Expectation[f(x; \parameter)\1[z=0]]$ during the last step of optimization. We construct a group-dependent factor $\alpha_{z} \coloneqq \1[z=1] - \1[z=0]$ by assigning $\alpha_0 = -1$ and $\alpha_1 = +1$. Then we can eliminate the absolute value notation in the $\hat{\constraint}(f)$ as follows:
\begin{align}\label{eq:decompose-relaxation}
    \hat{\constraint}(f) & = \frac{1}{n} \sum_{i=1}^n f(x_i;\parameter) (\1[z_i=1] - \1[z_i=0]) \nonumber \\
    & = \frac{1}{n} \sum_{i=1}^n \alpha_{z_i} f(x_i;\parameter)
\end{align}
Equation \ref{eq:decompose-relaxation} is saying the relaxed demographic parity constraint satisfies the decomposability assumption with
\begin{align}
    \hat{\constraint}(f, i) = \alpha_{z_i} f(x_i; \parameter)
\end{align}
Applying Lemma \ref{lem:constrained-influence}, we obtain the influence of demographic parity constraint for a training example $i$ on the target example $j$
\begin{align}\label{eq:relaxation-influence}
    S_{\textsf{DP}}(i, j) = \lambda\frac{\eta}{n} \alpha_{z_i} \Theta(x_i, x_j; \parameter_0) 
\end{align}
The above derivation presumes that the quantity inside the absolute value notation in Equation \ref{eq:relax-dp} is non-negative. In the opposite scenario where group $z=0$ is more favorable, we only need to reverse the sign of $\alpha_z$ to apply Equation \ref{eq:relaxation-influence}. We note that in some cases, the sign of the quantity will flip after one-step optimization, violating this assumption.

\paragraph{Equality of Opportunity.} For ease of notation, we define the utilities of True Positive Rate (TPR) and False Positive Rate (FPR) for each group $z \in \gZ$ as
\begin{align}
    \TPR_z & \coloneqq \Pr(f(x) \geq 0 \mid z = z, y = 1) \\
    \FPR_z & \coloneqq \Pr(f(x) \geq 0 \mid z = z, y = 0)
\end{align}
For the equal TPR measure $\psi(f) = |\TPR_1 - \TPR_0|$, we may relax it by
\begin{equation}
    \begin{multlined}
    \constraint(f) = \left|\Expectation [f(x; \parameter)\cdot\1[z = 1, y=1]]\right. \\ \qquad \left. - \Expectation[f(x; \parameter)\cdot\1[z=0, y=1]]\right|
    \end{multlined}
\end{equation}
Without loss of generality, we assume that the group $z=1$ has a higher utility such that the quantity within the absolute value notation is positive. We may construct the group-dependent factor $$\alpha_{z, y} \coloneqq \1[z=1, y=1] - \1[z=0, y=1]$$ by assigning $\alpha_{0, 1} = -1$, $\alpha_{1, 1} = +1$, and $\alpha_{z, -1} = 0$ for $z \in \{0, 1\}$. Then we may decompose $\hat{\constraint}(f)$ into
\begin{align}
    \hat{\constraint}(f) & = \frac{1}{n} \sum_{i=1}^n f(x_i; \parameter) \1[y_i=1](\1[z_i=1] - \1[z_i=0])
\end{align}
The above equation satisfies the decomposability assumption with $\hat{\constraint}(f, i) = \alpha_{z_i, y_i} f(x_i; \parameter)$. Applying Lemma \ref{lem:constrained-influence} again, we obtain the influence of equal TPR constraint
\begin{align}
    S_{\textsf{TPR}}(i, j) = \lambda\frac{\eta}{n} \alpha_{z_i, y_i} \Theta(x_i, x_j; \parameter_0)
\end{align}
For the equal FPR measure $\psi(f) = |\FPR_1 - \FPR_0|$, we may relax it by
\begin{equation}
\begin{multlined}
    \phi(f) = \left|\Expectation [f(x; \parameter)\cdot\1[z = 1, y = -1]]\right. \\
    \qquad \left.- \Expectation[f(x; \parameter)\cdot\1[z=0, y=-1]]\right|
\end{multlined}
\end{equation}
Likewise, we still assume the group $z=1$ has a higher utility. We construct the factor $$\tilde{\alpha}_{z, y} \coloneqq \1[z=1, y=-1] - \1[z=0, y=-1]$$ by assigning $\tilde{\alpha}_{0, -1} = -1$, $\tilde{\alpha}_{1, -1} = +1$, and $\tilde{\alpha}_{z, +1} = 0$ for $z \in \{0, 1\}$. Following the similar deduction, we may verify the relaxed equal FPR measure satisfies the decomposability assumption. Then we can obtain the influence of equal FPR constraint as
\begin{align}
     S_{\textsf{FPR}}(i, j) = \lambda \frac{\eta}{n} \tilde{\alpha}_{z_i, y_i} \Theta(x_i, x_j; \parameter_0) 
\end{align}
In the opposite scenario when group $z=0$ has a higher utility of either TPR or FPR, we may reverse the sign of $\alpha_{z, y}$ or $\tilde{\alpha}_{z, y}$, respectively. Finally, imposing equal odds constraint is identical to imposing equal TPR and equal FPR simultaneously, implying the following equality holds:
\begin{align}
    S_{\textsf{EO}} = S_{\textsf{TPR}} + S_{\textsf{FPR}}
\end{align}
\begin{corollary}\label{cor:DP-EO-relation}
    When one group has higher utilities (TPR and FPR) than the other group, the influence of imposing equal odds $S_\textsf{EO}(i, j)$ is equivalent to that of imposing demographic parity $S_\textsf{DP}(i,j)$.
\end{corollary}

\subsection{Covariance as Constraint}
Another common approach is to reduce the covariance between the group membership $z$ and the encoded feature $f(x;\parameter)$ \cite{Zafar2017FairnessCM,pmlr-v65-woodworth17a}. Formally, the covariance is defined by
\begin{align}
    \Covariance(z, f(x)) = \Expectation[z \cdot f(x;\parameter)] - \Expectation[z]\cdot \Expectation[f(x;\parameter)]
\end{align}
Then the empirical fairness measure is the absolute value of covariance
\begin{align}\label{eq:covariance-constraint}
    \hat{\phi}(f) = \left|\frac{1}{n}\sum_{i=1}^n z_i f(x_i;\parameter) - (\frac{1}{n} \sum_{i=1}^n z_i)\cdot (\frac{1}{n} \sum_{i=1}^n f(x_i;\parameter))\right|
\end{align}
Since we can observe the whole training set, the mean value of group membership can be calculated by $\bar{z} = \frac{1}{n} \sum_{i=1}^n z_i$. As a result, we can decompose the covariance as follows:
\begin{align}
    \hat{\phi}(f) & = |\frac{1}{n}\sum_{i=1}^n (z_i - \bar{z}) f(x_i; \parameter)| \nonumber \\
    & = \frac{1}{n} \sum_{i=1}^n \beta_i (z_i - \bar{z}) f(x_i; \parameter)
\end{align}
where $\beta_i \in \{-1, +1\}$ is an instance-dependent parameter. Then the covariance constraint satisfies the decomposability assumption by taking
\begin{align}
    \hat{\phi}(f, i) = \beta_i (z_i - \bar{z}) f(x_i; \parameter), \quad \beta_i \in \{-1, +1\}.
\end{align}
Finally, the influence score induced by the covariance constraint in Equation \ref{eq:covariance-constraint} is
\begin{equation}\label{eq:covariance-influence}
    S_{\textsf{cov}}(i, j) = \lambda \frac{\eta}{n} \beta_i (z_i - \bar{z}) \Theta(x_i, x_j ; \parameter_0)
\end{equation}
In this kernelized expression, the pairwise influence score is neatly represented as NTK scaled by an instance weight $\beta_i (z_i - \bar{z})$. 

\paragraph{Connection to Relaxed Constraint.} We may connect the influence function of the covariance approach to that of the relaxation approach in a popular situation where there are only two sensitive groups.
\begin{corollary}\label{cor:relaxation-covariance-connection}
    When sensitive attribute $z$ is binary, the influence score of covariance is half of the influence of relaxed demographic parity.
    \[
        |\gZ| = 2 \implies S_\textsf{cov}(i,j) = \frac{1}{2} S_\textsf{DP}(i,j)
    \]
\end{corollary}

\begin{figure*}[!ht]
    \centering
    \subfigure{
        \includegraphics[width=0.4\linewidth]{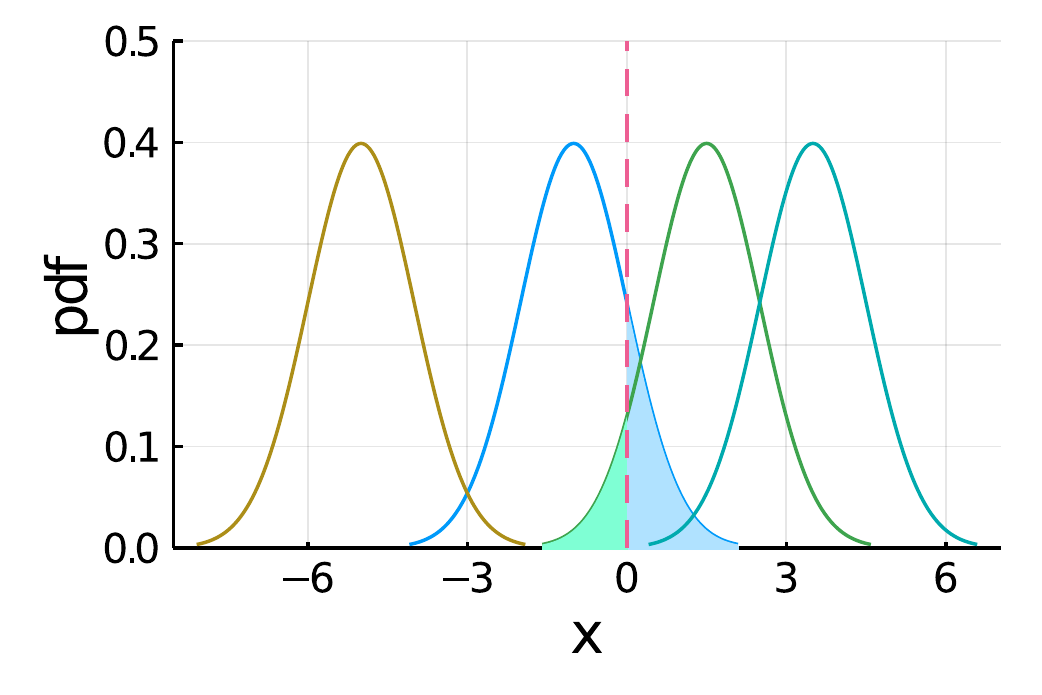}
    }
    \subfigure{
        \includegraphics[width=0.54\linewidth]{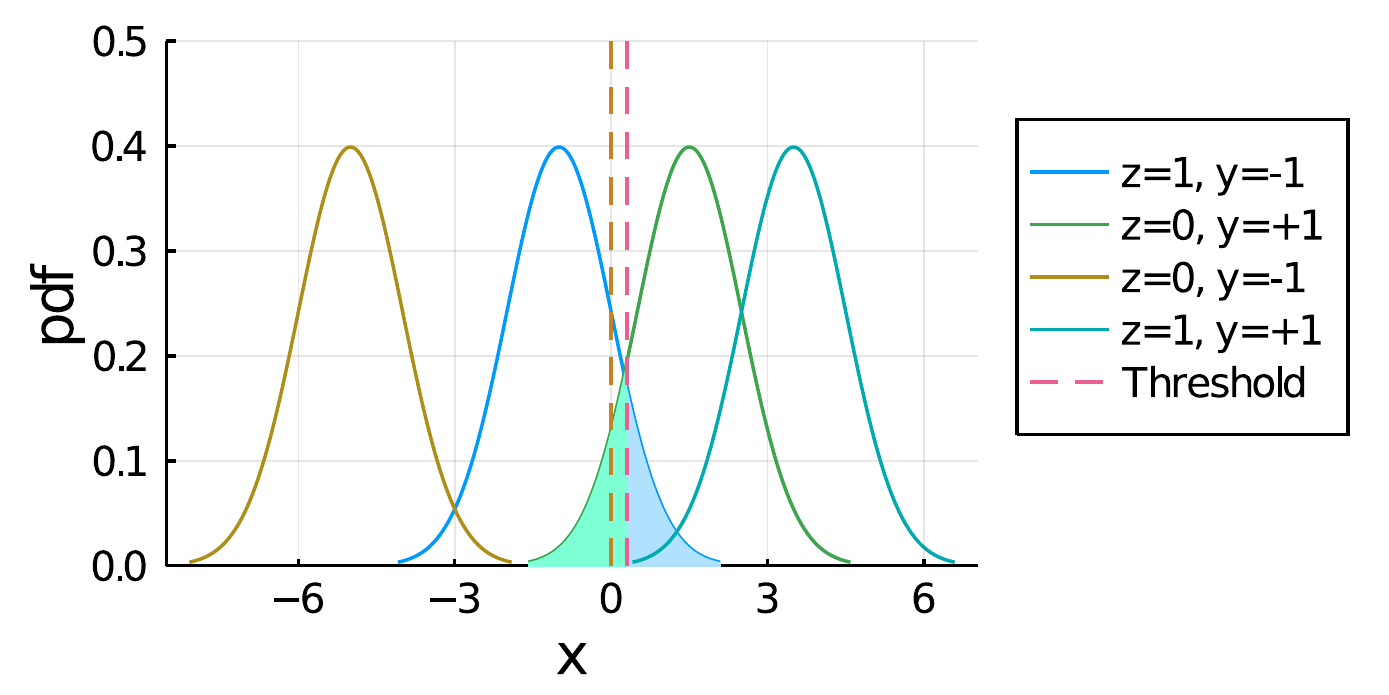}
    }
    \caption{A toy example to interpret the influence scores of fairness. \textbf{Left:} The optimal classifier is $\1[x \geq 0]$. Four curves in different colors represent the distributions for each $(z,y)$ combination. The blue area and green area represent the violation of demographic parity. The data examples with low influence scores of fairness constraints are around $x=0$. \textbf{Right:} When we down-weight the examples around $x=0$, the optimal classifier will be perturbed towards right. Since the sum of blue area and green area decreases, the violation of demographic parity is mitigated.}
    \label{fig:prune_example}
\end{figure*}

\subsection{Information Theoretic Algorithms}
The demographic parity constraint can be interpreted as the independence of prediction $f(x)$ and group membership $z$. Denoted by $I(f(x); z)$ the mutual information between $f(x)$ and $z$, the independence condition $f(x) \independent z$ implies $I(f(x); z) = 0$. In consequence, a number of algorithms \cite{Song2019LearningCF,Gupta2021ControllableGF,Baharlouei2020Renyi} propose to adopt the bounds of mutual information $I(f(x); z)$ as the empirical fairness measure. We consider approximating mutual information by MINE \cite{pmlr-v80-belghazi18a,Oord2018RepresentationLW} as an example.
\begin{align}
    \hat{\constraint}(f) = \frac{1}{n} \sum_{i=1}^n \log \frac{\exp g(f(x_i), z_i)}{\frac{1}{n} \sum_{k=1}^{n} \exp g(f(x_i), z_k)}
\end{align}
where the function $g(\cdot, \cdot)$ is parameterized by a neural network. In this case, $\hat{\constraint}(f)$ satisfies the decomposability assumption by straightly taking
\begin{align}\label{eq:decompose-InfoNCE}
    \hat{\constraint}(f, i) = \log\frac{\exp g(f(x_i), z_i)}{\frac{1}{n} \sum_{k=1}^{n} \exp g(f(x_i), z_k)}
\end{align}
Although the denominator inside the logarithm in Equation \ref{eq:decompose-InfoNCE} contains the sum over all the $z_k$ in the training set, we can always calculated the sum when we know the prior distribution of the categorical variable $z$. Taking the derivative of $\hat{\constraint}(f, i)$, the influence of MINE constraint is given by
\begin{equation}\label{eq:influence-InfoNCE}
\begin{aligned}
    S_\textsf{MINE}(i, j) & = \lambda\frac{\eta}{n} \Theta(x_i, x_j; \parameter_0)\cdot \at{\pdv{\mathbf{G}}{w}}{w=f(x_i; \parameter)} \\ 
    \text{where}\quad \mathbf{G} & = g(w, z_i) - \frac{1}{n}\sum_{k=1}^n g(w, z_k)
\end{aligned}
\end{equation}

\paragraph{Connection to Covariance.} In a special case when $g(f(x; \parameter), z) = zf(x;\parameter)$, we have $\partial_f g(f(x; \parameter), z) = z$. Substitute the partial derivative back into Equation \ref{eq:influence-InfoNCE}, the influence of MINE reduces to the influence of covariance measure $\lambda \frac{\eta}{n} \alpha_i (z_i - \bar{z}) \Theta(x_i, x_j ; \parameter_0)$. However, it is very likely that the influence scores of MINE and covariance are much different, due to the fact that the unknown function $g(f(x), z)$ is parameterized by neural networks in more generic applications.

%% file: Section/GeneralizationBound.tex
\section{Estimating the Aggregated Influence Score}
In this section, we intend to discuss the expected influence of a training example on the whole data distribution. We will focus on the changes of empirical fairness constraints $\hat{\phi}(f)$ when a data point $(x_i, y_i, z_i)$ is excluded in the training set or not. Suppose that $\hat{\phi}(f)$ satisfies the decomposability assumption. We define the realized influence score of a training example $i$ aggregated over the whole data distribution $\data$ as
\begin{align}
    \mathcal{S}(i) := \int_{(x_j, y_j, z_j) \in \data} \pdv{\hat{\phi}(f, j)}{f} S (i, j)\,d \Pr(x_j, y_j, z_j)
\end{align}
$\mathcal{S}(i)$ takes into account the change on $\hat{\phi}(f)$ by applying the first-order approximation again for each test point $j$. In practice, the model $f$ can only observe finite data examples in $D$ that are drawn from the underlying distribution $\data$. We estimate the influence score of a training example $i$ over the training set $D$.
\begin{align}\label{eq:empirical-influence-score-fairness}
    S(i) := \frac{1}{n}\sum_{i=1}^n \pdv{\hat{\phi}(f, j)}{f} S(i, j)
\end{align} 
We wonder how the measure of $S(i)$ deviates from $\mathcal{S}(i)$.
\begin{theorem}[Generalization Bound]\label{thm:generalization-bound}
With probability at least $1 - \epsilon$,
\begin{align}
    \mathcal{S}(i) - S(i) \leq \mathcal{O}\left(\sqrt{\frac{\log \frac{1}{\epsilon}}{2n}}\right)
\end{align}
\end{theorem}

\paragraph{Interpreting Influence Scores on Synthetic Data.}
We consider a synthetic example as visualized in Figure \ref{fig:prune_example} to illustrate why our influence scores help with identifying instances that affect the fairness. We assume that the individual examples are independently drawn from an underlying normal distribution corresponding to the label $y \in \{-1, +1\}$ and group membership $z \in \{0, 1\}$, i.e., $x_{z, y} \sim N(\mu_{z, y}, \sigma)$. We assume that $\mu_{0, -1} < \mu_{1, -1} < 0 < \mu_{0, +1} < \mu_{1, +1}$. Suppose that we train a linear model $f(x) = w\cdot x + b$, and the obtained classifier is $\1[f(x) >= 0]$ which reduces to $\1[x >= 0]$ for our toy example. Then we have the following proposition:
\begin{proposition}\label{prop:example}
In our considered setting, if we down-weight the training examples with smaller absolute fairness influence scores, the model will tend to mitigate the violation of demographic parity.
\end{proposition}
Proposition \ref{prop:example} informs us that we can mitigate the unfairness by up-weighting the data instances with higher influence scores, or equivalently by removing some low-influence training points. 

%% file: Section/Experiment.tex
\section{Evaluations}\label{sec:evaluation}
In this section, we examine the influence score subject to parity constraints on three different application domains: tabular data, images and natural language.
\begin{figure*}[!thb]
    \centering
    \subfigure[]{
        \includegraphics[width=0.3\linewidth]{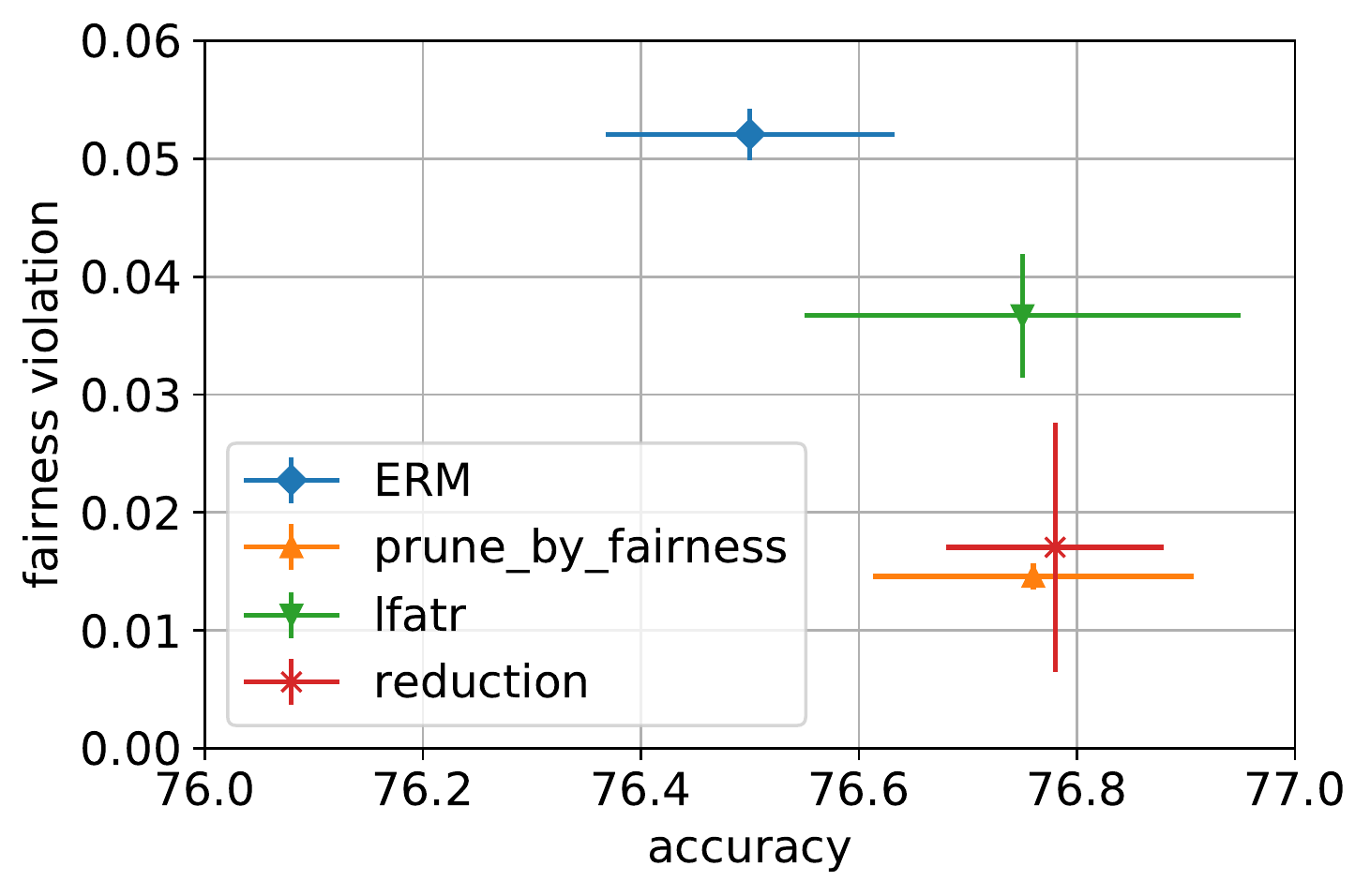}
        \label{fig:adult_result_a}
    }
    \hfill
    \subfigure[]{
        \includegraphics[width=0.3\linewidth]{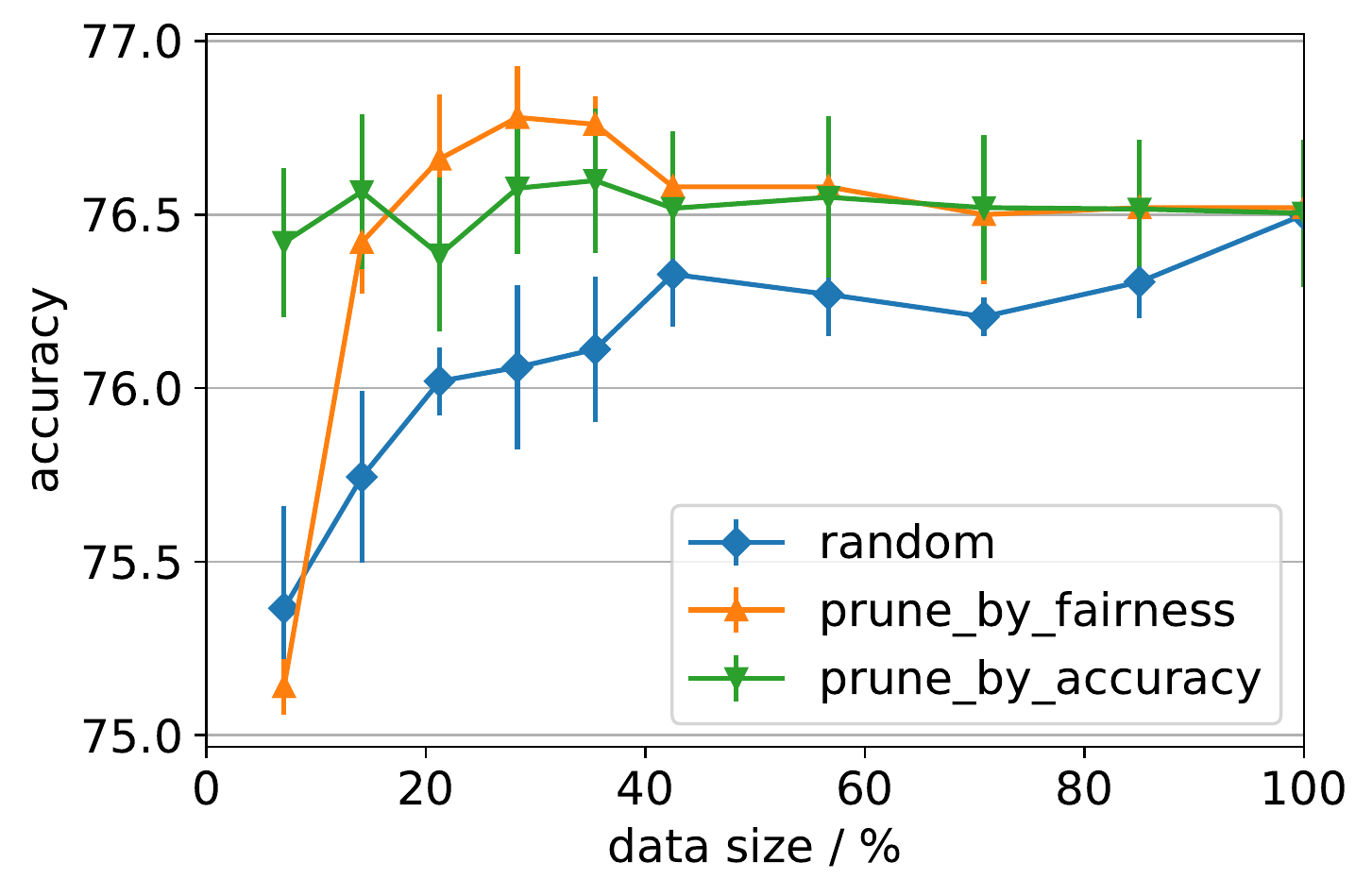}
        \label{fig:adult_result_b}
    }
    \hfill
    \subfigure[]{
        \includegraphics[width=0.3\linewidth]{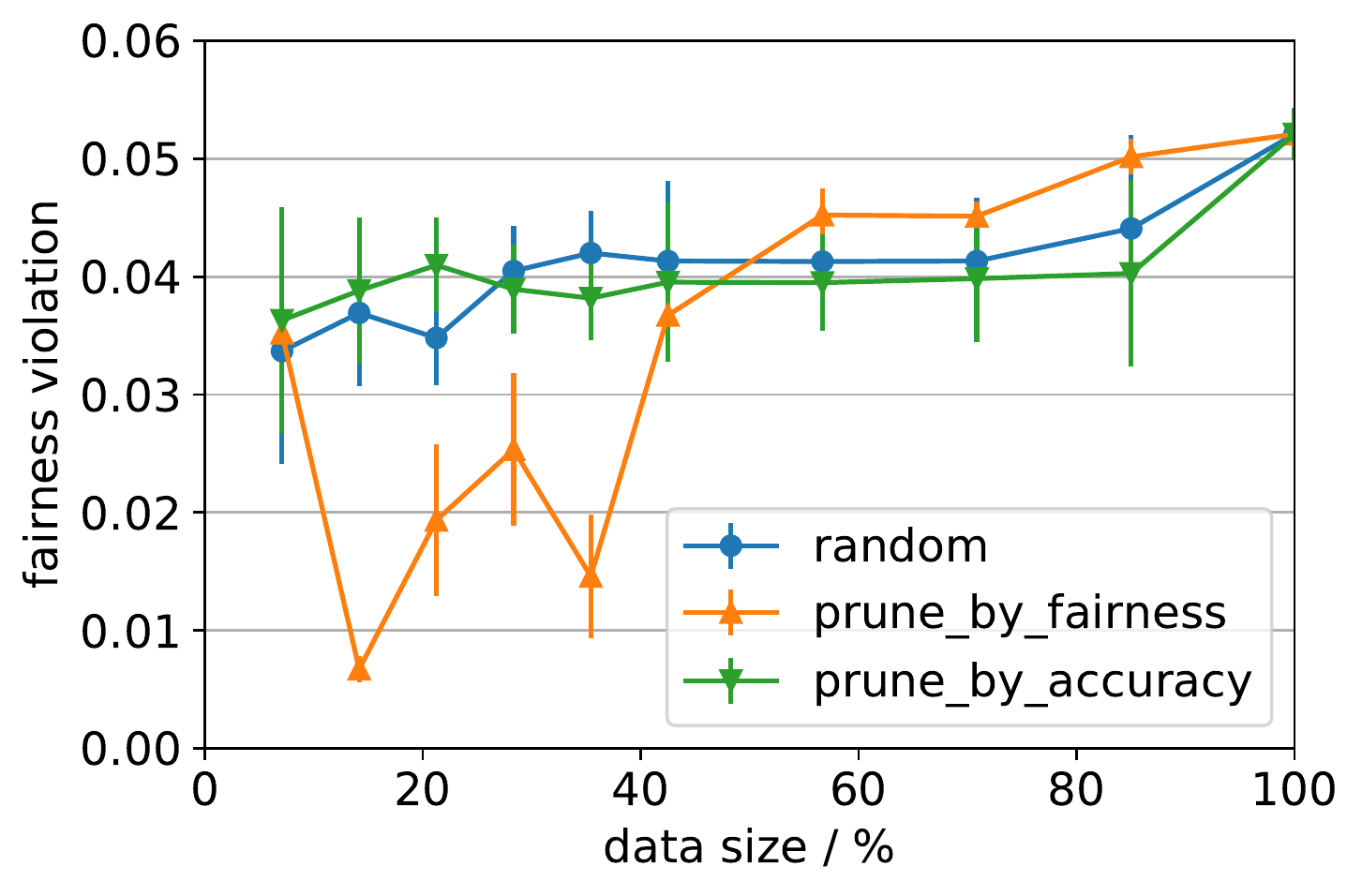}
        \label{fig:adult_result_c}
    }
    \caption{Results on Adult dataset. Figure \ref{fig:adult_result_a}: we benchmark the fairness and accuracy metrics for the baselines. Figure \ref{fig:adult_result_b} and Figure \ref{fig:adult_result_c}: we compare how the proportion of unpruned training data affect the accuracy and fairness violation, respectively.}
    \label{fig:adult_result}
\end{figure*}

\begin{figure*}[!thb]
    \centering
    \subfigure[]{
        \includegraphics[width=0.3\linewidth]{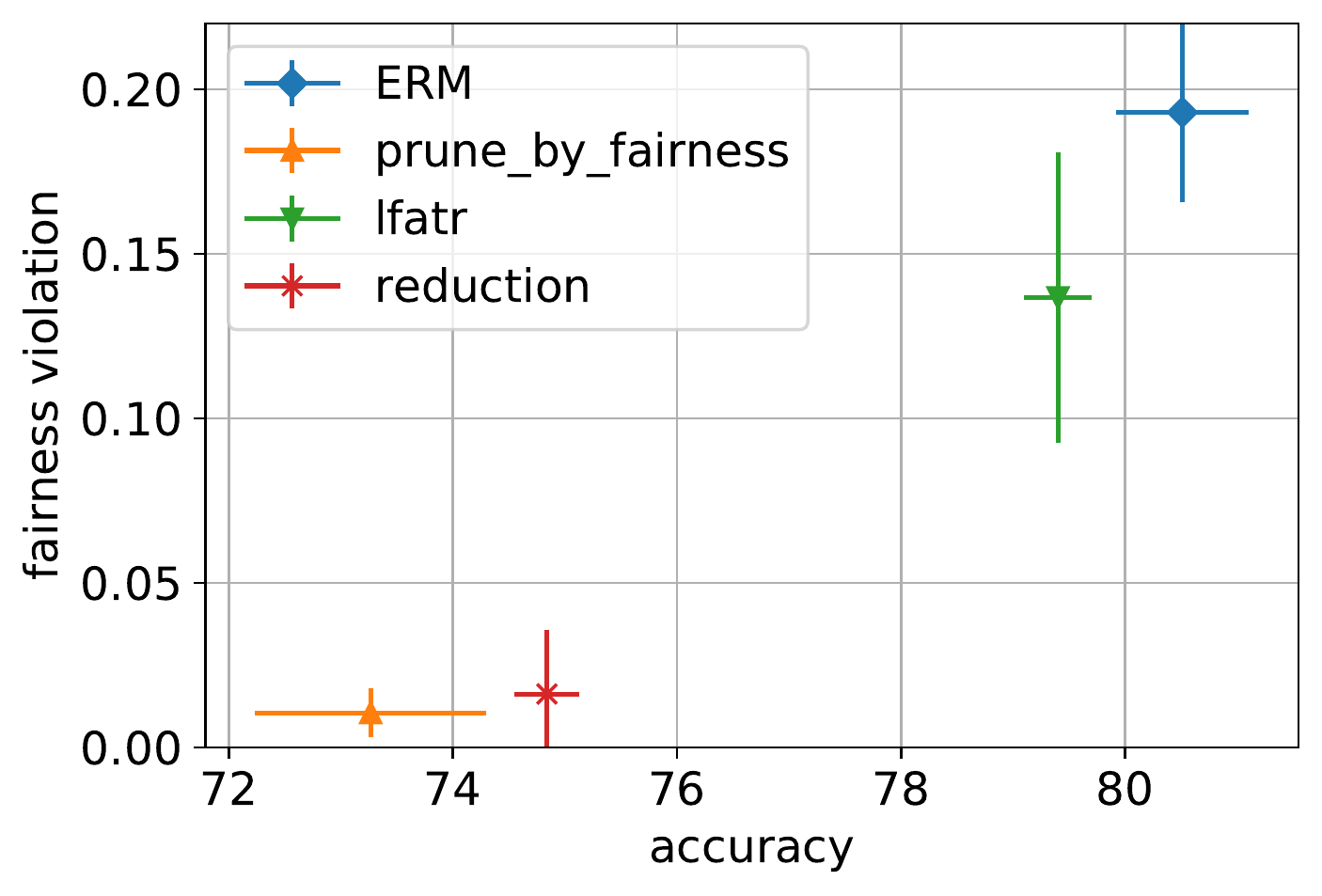}
        \label{fig:celeba_result_a}
    }
    \hfill
    \subfigure[]{
        \includegraphics[width=0.3\linewidth]{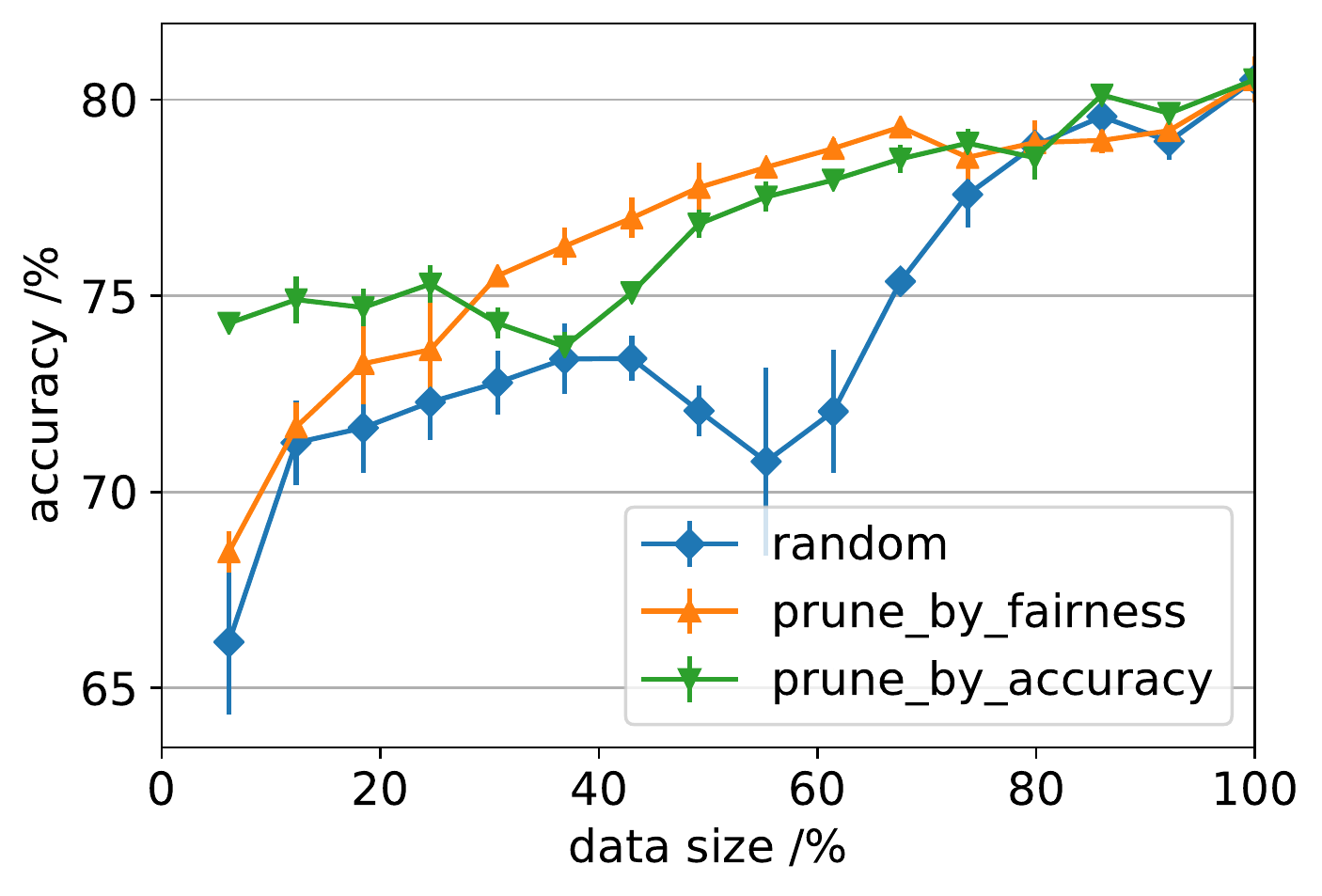}
        \label{fig:celeba_result_b}
    }
    \hfill
    \subfigure[]{
        \includegraphics[width=0.3\linewidth]{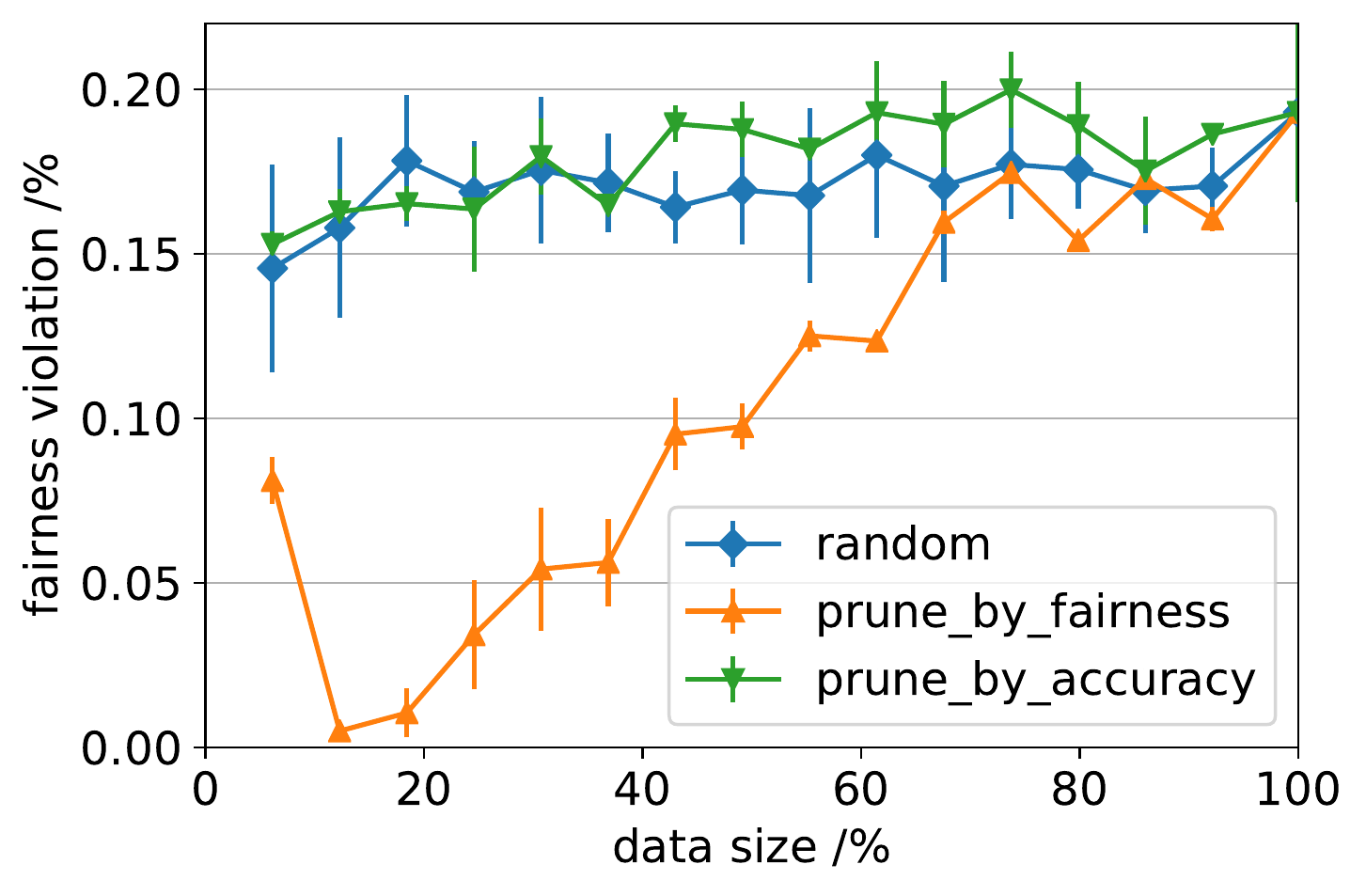}
        \label{fig:celeba_result_c}
    }
    \caption{Results on CelebA dataset. Figure \ref{fig:celeba_result_a}: we benchmark the fairness and accuracy metrics for the baselines. Figure \ref{fig:celeba_result_b} and Figure \ref{fig:celeba_result_c}: we compare how the proportion of unpruned training data affect the accuracy and fairness violation, respectively.}
    \label{fig:celeba_result}
\end{figure*}

\begin{figure*}[!thb]
    \centering
    \subfigure[]{
        \includegraphics[width=0.3\linewidth]{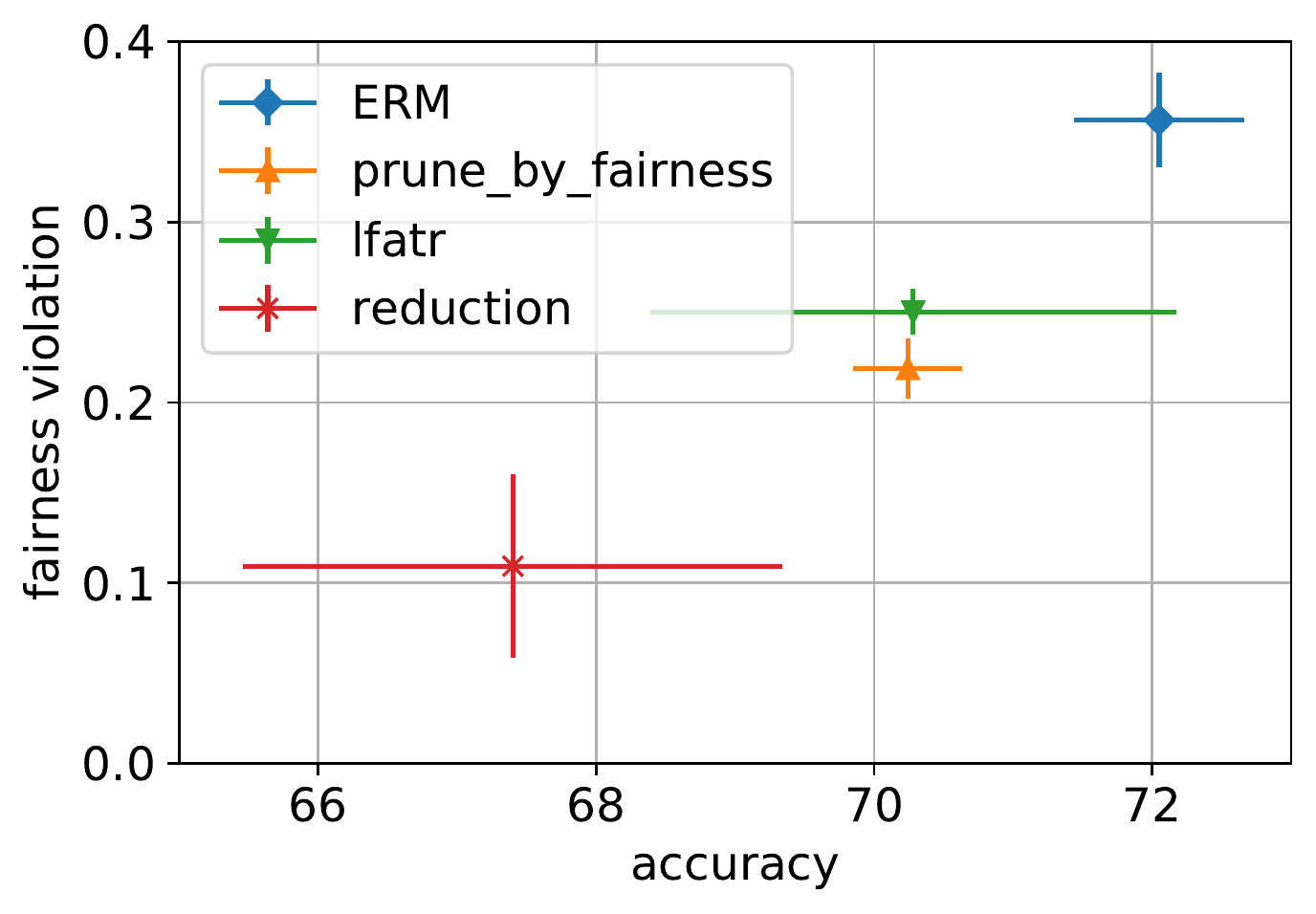}
        \label{fig:jigsaw_result_a}
    }
    \hfill
    \subfigure[]{
        \includegraphics[width=0.3\linewidth]{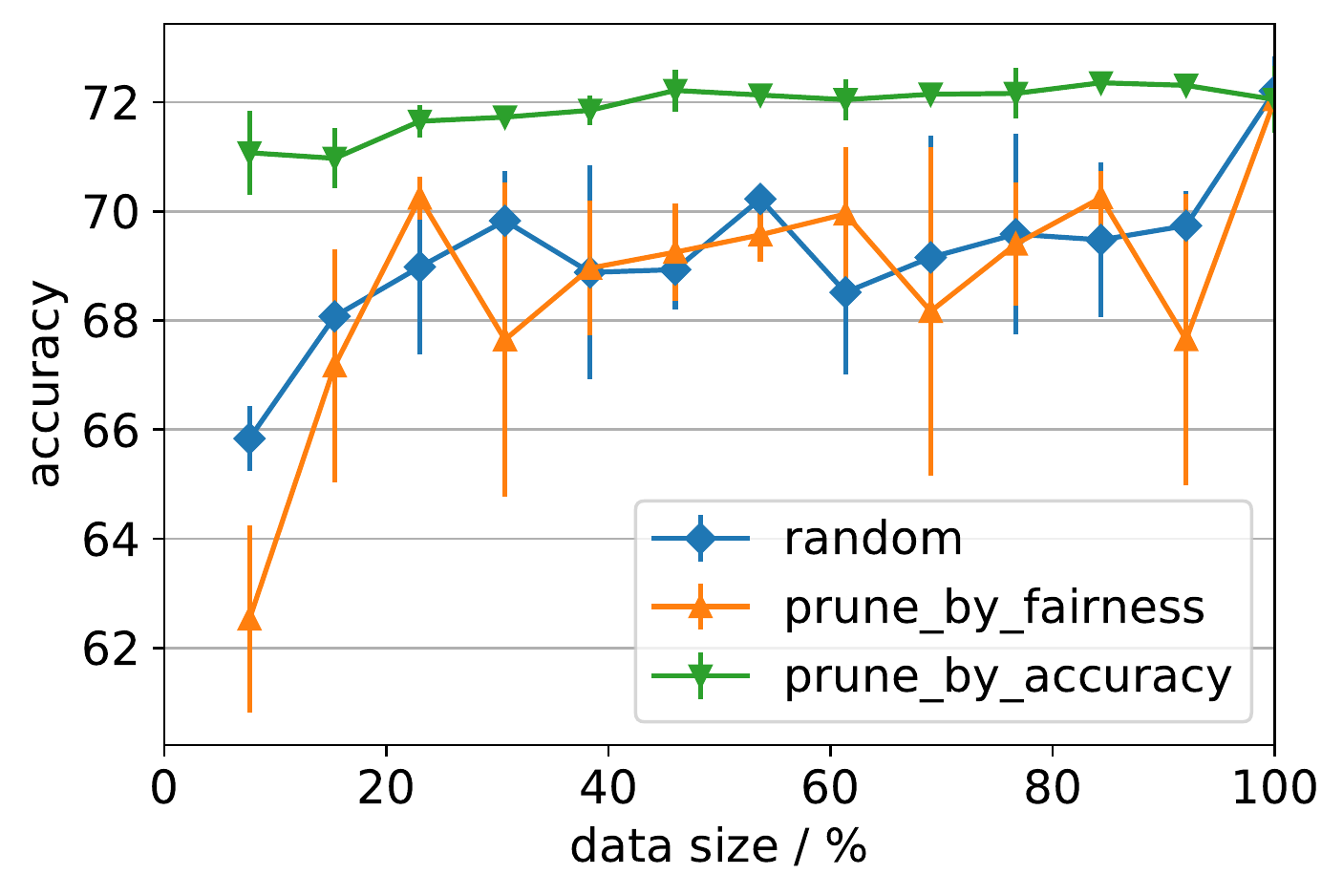}
        \label{fig:jigsaw_result_b}
    }
    \hfill
    \subfigure[]{
        \includegraphics[width=0.3\linewidth]{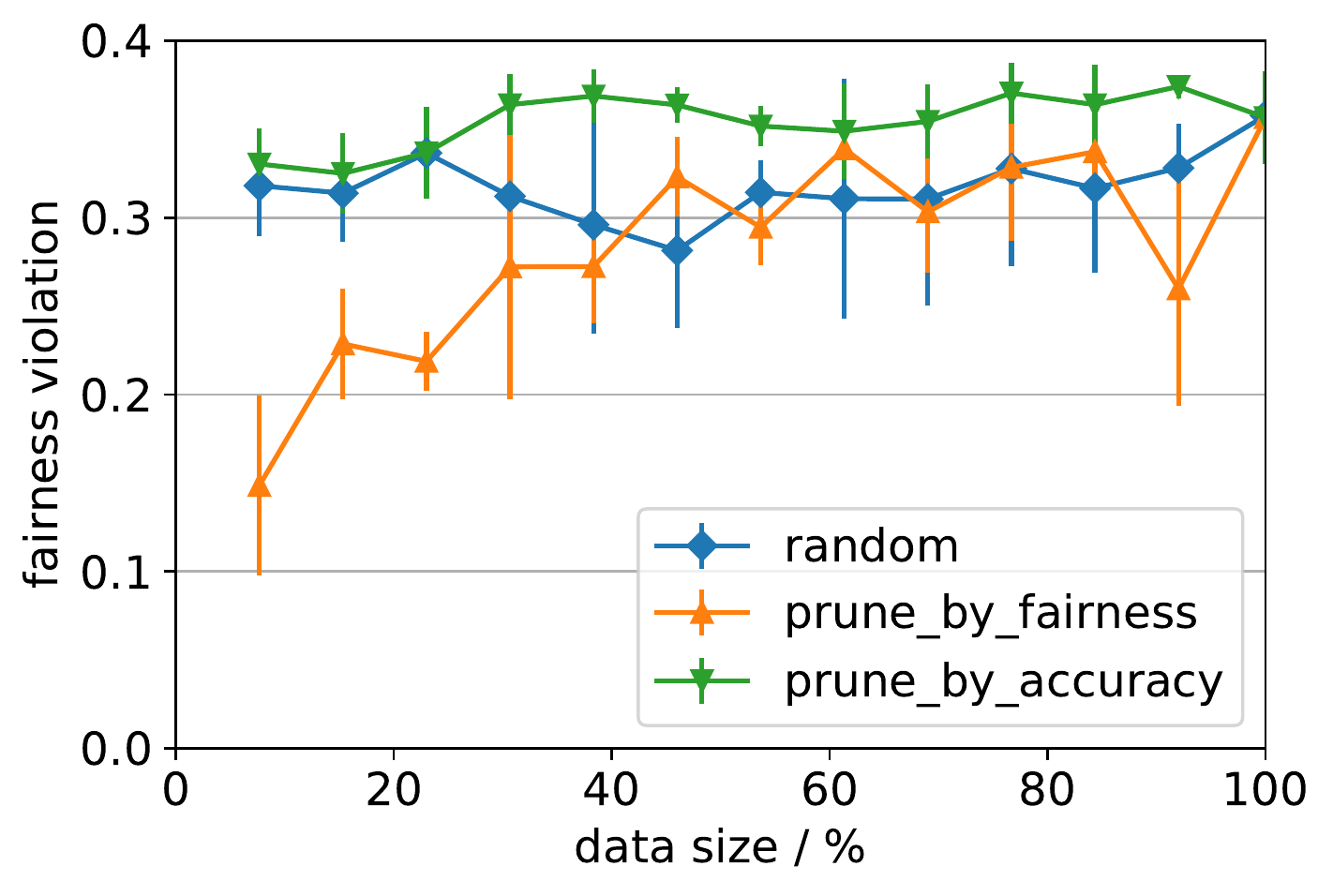}
        \label{fig:jigsaw_result_c}
    }
    \caption{Results on Jigsaw dataset. Figure \ref{fig:jigsaw_result_a}: we benchmark the fairness and accuracy metrics for the baselines. Figure \ref{fig:jigsaw_result_b} and Figure \ref{fig:jigsaw_result_c}: we compare how the proportion of unpruned training data affect the accuracy and fairness violation, respectively.}
    \label{fig:jigsaw_result}
\end{figure*}

\subsection{Setup}
We adopt the evaluation protocol that has been widely used by the previous papers on interpreting the impact of data examples. The basic idea is to train a model on a subset of training data by removing the less influential examples \cite{paul2021deep}. In particular, we assess the performance of three data prune strategies: (1) random, which randomly selects a subset of training examples; (2) prune by fairness, which removes the data examples in the ascending order of the absolute values of aggregated influence scores subject to fairness as described in Equation \ref{eq:empirical-influence-score-fairness}; (3) prune by accuracy, which removes the data examples in the descending order of absolute influence scores in terms of the loss. The influence score is equivalent to the first order approximate proposed in \cite{tracin}. For the three strategies above, a model pre-trained on the whole training set will be used to estimate the influence scores of training examples with the direct application of Equation \ref{eq:empirical-influence-score-fairness}. We then execute the data prune procedure and impose the relaxed demographic parity in Equation \ref{eq:relax-dp} to train a fair model.

We compare the performance of pruning by influence scores with following optimization algorithms that regularize the constraint:
\begin{itemize}
    \item ERM, which trains the model directly without imposing fairness constraints.
    \item lfatr \cite{DBLP:journals/corr/abs-1802-06309}, which regularizes the model with relaxed constraint as given in Equation \ref{eq:decompose-relaxation}.
    \item reduction \cite{fairlearn}, which reduces the constrained optimization to a cost-sensitive learning problem.
\end{itemize}
We used the Adam optimizer with a learning rate of 0.001 to train all the models. We used $\gamma=1$ for models requiring the regularizer parameter of fairness constraints. Any other hyperparameters keep the same among the compared methods. We report two metrics: the accuracy evaluated on test set and the difference of acceptance rates between groups as fairness violation. We defer more experimental details to Appendix \ref{appendix:experiments}.

\subsection{Result on Tabular Data}
Firstly, we work with multi-layer perceptron (MLP) trained on the Adult dataset \cite{dua2017uci}. We select sex, including female and male, as the sensitive attribute. We resample the dataset to balance the class and group membership. The MLP model is a two-layer ReLU network with hidden size 64. We train the model 5 times with different random seeds to report the mean and standard deviation of accuracy and fairness metrics. In each trial, the dataset is randomly split into a training and a test set in a ratio of 80 to 20.
We compare the performance of prune by fairness in Figure \ref{fig:adult_result_a} and find it has a similar fairness-accuracy trade-off with the reduction approach. To gain further insights, we plot how the size of pruned training examples affects the accuracy and fairness metrics for three prune strategies in Figure \ref{fig:adult_result_b} and Figure \ref{fig:adult_result_c}. Not surprisingly, the random baseline remains a high fairness violation with a large accuracy drop when the data size decreases. In contrast, prune by fairness has a similar accuracy with pruning by accuracy when the data size is greater than $20\%$ and mitigates the fairness violation by a large margin when the data size is less than $40\%$. We also notice that prune by fairness anomalously has a high fairness violation when the data size is less than $10\%$. We conjecture such a small size of training data does not contain sufficient information, leading to the significant performance degradation. 
These observations suggest that we may obtain the best trade-off with a subset of only $20\%$--$40\%$ of training data.

\subsection{Result on Images}
Next, we train a ResNet-18 network \cite{DBLP:journals/corr/HeZRS15} on the CelebA face attribute dataset \cite{CelebA}. We select smiling as binary classification target and gender as the sensitive attribute. Figure \ref{fig:celeba_result_a} shows the trade-off between accuracy and fairness violation for each baseline method. We explore how the size of pruned training examples affects the accuracy and fairness metrics in Figure \ref{fig:celeba_result_b} and Figure \ref{fig:celeba_result_c}. Again, the accuracy of prune by fairness has a similar trend with that of prune by accuracy when data size is larger than $20\%$, but drops strikingly with much smaller data size. On the other hand, prune by fairness mitigates the fairness violation straightly when data size decreases.

\subsection{Result on Natural Language}
Lastly, we consider Jigsaw Comment Toxicity Classification \cite{jigsaw} with text data. We select race as the sensitive attribute in our evaluation. We use pre-trained BERT \cite{devlin-etal-2019-bert} to encode each raw comment text into a $768$-dimensional textual representation vector and train a two-layer neural network to perform classification. We report the experimental result in Figure \ref{fig:jigsaw_result}. Figure \ref{fig:jigsaw_result_a} shows that prune by fairness has a mimic performance with lfatr while preserving smaller standard deviation. Figure \ref{fig:jigsaw_result_b} shows that prune by accuracy keeps a relatively high accuracy when a large subset of training examples are removed. In comparison, both prune by fairness and random prune failed to make informative prediction when the data size is below $20\%$. Figure \ref{fig:jigsaw_result_c} implies that prune by fairness is capable of mitigating bias. This result cautions that we need to carefully account for the price of a fair classifier, particularly in this application domain.

%% file: Section/Discussion.tex
\section{Conclusion}
In this work, we have characterized the influence function subject to fairness constraints, which measures the change of model prediction on a target test point when a counterfactual training example is removed. We hope this work can inspire meaningful discussion regarding the impact of fairness constraints on individual examples. We propose exploring reasonable interpretations along this direction in real-world studies for future work.

%% file: Section/Appendix.tex
\section{Impact of Fairness Constraints on Loss}\label{appendix:influence-on-loss}
We may also track the change in the test loss for a particular test point $x_j$ when training on a counterfactual instance $x_i$. Analogous to Equation \ref{defi:influence-function}, we may define the influence function in terms of the change of loss by
\begin{equation}
    \infl_\ell (D, i, j) \coloneqq \ell(f^{D/\{i\}}(x_j), y_j) -  \ell(f^{D}(x_j;\parameter, y_j)
\end{equation}
In the same way, we may approximate the influence with Taylor expansion again,
\begin{align}
    \infl_{\ell} (D, i, j) & = \ell(f(x_j;\parameter, y_j) - \ell(f(x_j; \parameter_0), y_j) \nonumber \tag{by Definition}\\
    & \approx \at{\frac{\partial\ell(w, y_j)}{\partial w}}{w = f(x_j; \parameter)} \left(f(x_j;\parameter) - f(x_j;\parameter_0)\right) \nonumber \tag{by first-order Taylor expansion}\\
    & \approx \frac{\eta}{n} \at{\frac{\partial\ell(w, y_j)}{\partial w}}{w = f(x_j; \parameter_0)} \Theta(x_i, x_j; \parameter_0) \at{\frac{\partial\ell(w, y_i)}{\partial w}}{w = f(x_i; \parameter_0)} \tag{by substituting Equation \ref{eq:unconstrained-influence}}
\end{align}

In more complicated situation where the fairness constraints are regularized, the influence function would be
\begin{align}\label{eq:constrained-influence-loss}
    \infl_{\ell} (D, i, j) & = \ell(f(x_j;\parameter, y_j) - \ell(f(x_j; \parameter_0), y_j) \nonumber \\
    & \approx \at{\frac{\partial\ell(w, y_j)}{\partial w}}{w = f(x_j; \parameter_0)} (f(x_j;\parameter) - f(x_j;\parameter_0)) \nonumber \\
    & \approx \frac{\eta}{n} \at{\frac{\partial\ell(w, y_j)}{\partial w}}{w = f(x_j; \parameter_0)} \Theta(x_i, x_j; \parameter_0)  \left(\at{\frac{\partial\ell(w, y_i)}{\partial w}}{w = f(x_i; \parameter_0)} + \lambda \at{\frac{\partial\hat{\constraint}(f, i)}{\partial f}}{f(x_i; \parameter_0)}\right)
\end{align}
Equation \ref{eq:constrained-influence-loss} implies the intrinsic tension between accuracy and fairness --- when the sign of $\pdv*{\ell(f(x_i), y_i)}{f}$ are opposed to that of $\pdv*{\hat{\constraint}(f, i)}{f}$, the influence of parity constraint will contradict with that of loss.

\section{Omitted Proofs}\label{appendix:omitted-proof}
\textbf{Proof of Corollary \ref{cor:DP-EO-relation}:}
    \begin{proof}
        Without loss of generality, we assume group $z=1$ has higher utilities than group $z=0$, i.e.,
       \begin{align*}
            \Expectation [f(x; \parameter)\1[z = 1, y = 1]] & \geq \Expectation[f(x; \parameter)\1[z=0, y=1]] \\
            \Expectation [f(x; \parameter)\1[z = 1, y = 0]] & \geq \Expectation[f(x; \parameter)\1[z=0, y=0]]
        \end{align*}  
        Equal odds indicates equal TPR and equal FPR constraints will be imposed simultaneously. Thereby
        \begin{align*}
            S_{\textsf{EO}} & = S_{\textsf{TPR}} + S_{\textsf{FPR}} \\
            & = \lambda\frac{\eta}{n} \alpha_{z_i, y_i} \Theta(x_i, x_j; \parameter_0)  + \lambda \frac{\eta}{n} \tilde{\alpha}_{z_i, y_i} \Theta(x_i, x_j; \parameter_0) \\
            & = \lambda\frac{\eta}{n} \alpha_{z_i} \Theta(x_i, x_j; \parameter_0)  \tag{by $\alpha_z = \alpha_{z, y} + \tilde{\alpha}_{z, y}$} \\
            & = S_{\textsf{DP}}
        \end{align*}
        The third equality is due to $\alpha_z = \1[z=1] - \1[z=0] = (\1[z=1, y=+1] + \1[z=1, y=-1]) - (\1[z=0, y=+1] + \1[z=0, y=-1]) = \alpha_{z, y} + \tilde{\alpha}_{z, y}$.
    \end{proof}

\textbf{Proof of Corollary \ref{cor:relaxation-covariance-connection}:}
\begin{proof}
        When there are only two groups, the covariance measure in Equation \ref{eq:covariance-constraint} reduces to
        \begin{align*}
            \hat{\phi}(f) = |\frac{1}{n} \sum_{i=1}^n (z_i - \frac{1}{2}) f(x_i; \parameter)|
        \end{align*}
        Again, we assume group $z=1$ is more favorable than group $z=0$ such that $\Expectation [f(x; \parameter)\1[z = 1]] \geq \Expectation[f(x; \parameter)\1[z=0]]$. Then we can rewrite the above equation as
        \begin{align*}
            \hat{\phi}(f) & = \frac{1}{n} \sum_{i=1}^n \frac{1}{2} f(x_i;\parameter) \1[z_i=1] - \frac{1}{n} \sum_{i=1}^n \frac{1}{2}f(x_i;\parameter) \1[z_i=0]) \geq 0
        \end{align*}
        The above $\hat{\phi}(f)$ is saying the covariance between $z$ and $f(x)$ is non-negative per se, so we do not need to take the absolute value of it. In other words, $\forall i, \beta_i = 1$. The final influence score of covariance thus becomes
        \[
            S_\textsf{cov}(i,j) = \lambda \frac{\eta}{2n} \Theta(x_i, x_j ; \parameter_0) (\1[z_i=1] - \1[z_i=0])
        \]
        Recall that $\alpha_i = \1[z_i=1] - \1[z_i=0]$, we conclude $S_\textsf{cov}(i,j) = \frac{1}{2} S_\textsf{DP}(i,j)$. We note that the connection builds upon the common assumption that group $z=1$ has a higher utility. We can reach the same conclusion in the symmetric situation where $\Expectation [f(x; \parameter)\1[z = 1]] < \Expectation[f(x; \parameter)\1[z=0]]$.

        We remark, the coefficient $\frac{1}{2}$ arises from encoding the categorical sensitive variable $z$ into $\{0, 1\}$ and does not have physical meanings. If $z$ is encoded by $\{-1, +1\}$ instead, the coefficient will be $1$ such that $ S_\textsf{DP} = S_\textsf{cov}$. This property suggests that the covariance is not a perfect measure of independence, and using mutual information is a more plausible approach.
    \end{proof}
    
\textbf{Proof of Theorem \ref{thm:generalization-bound}}
\begin{proof}
For any $t$ and any $\delta > 0$,
\begin{align*}
    \Pr(\mathcal{S}(f, j) - S(f, j) > \delta) & = \Pr(\exp {nt\left(\mathcal{S}(f, j) - S(f, j)\right)} > \exp{nt\delta}) \\
        & \leq \frac{\Expectation[\exp{nt(\mathcal{S}(f, j) - S(f, j))}]}{\exp{nt\delta}} \tag{by Markov's inequality} \\
        & \leq \exp{\frac{1}{8}nt^2C^2 - nt\delta} \tag{by Hoeffding's inequality}
\end{align*}
In above $C$ is some constant. Since $\frac{1}{8}nC^2t^2 - n\delta t$ is a quadratic function regarding $t$, we may minimize it by taking
\begin{align*}
    \pdv{t} (\frac{1}{8}nC^2t^2 - n\delta t)  = 0  \implies \frac{1}{4}nC^2 t - n \delta t  = 0 
\end{align*}
Solving the above equation, we know the quadratic function takes the minimum value at $t = \frac{\delta}{4C^2}$. Therefore,
\[
    \Pr\left(\mathcal{S}(f, j) - S(f, j) > \delta\right) \leq \exp{-\frac{2n\delta^2}{C^2}}
\]
Let $\epsilon = \exp{-\frac{2n\delta^2}{C^2}}$, we complete the proof by substituting $\delta$ with $\epsilon$
\[
    \Pr\left(\mathcal{S}(f, j) - S(f, j) > C \sqrt{\frac{\log \frac{1}{\epsilon}}{2n}}~\right) \leq \epsilon \implies \Pr\left(\mathcal{S}(f, j) - S(f, j) \leq C \sqrt{\frac{\log \frac{1}{\epsilon}}{2n}}~\right) > 1 - \epsilon
\]
\end{proof}

\textbf{Proof of Proposition \ref{prop:example}}
\begin{proof}
We visualize the considered example in Figure \ref{fig:prune_example}. The area in blue represents the false positive examples from group $z=1$, while the area in green represents the false negative examples from group $z=0$. The sum of blue area and green area is exactly representing the acceptance rate difference between group $z=1$ and $z=0$. 

Recall that the model is $f(x) = w\cdot x + b$, the influence function subject to relaxed fairness constraint can be computed by Equation \ref{eq:relaxation-influence}, i.e., $S(i, j) = k (z_i - \bar{z}) \cdot (x_i \cdot x_j + 1)$ where $k$ is a constant coefficient corresponding to learning rate $\eta$, data size $n$ and regularizer $\lambda$. For each individual example $x_i$, the overall influence score $S(i)$ consists of two components. The first component
\[
    \int_{x_j \in (-\infty, +\infty)} k(z_j - \bar{z}) \cdot (z_i - \bar{z}) \cdot x_i \cdot x_j \,d \Pr(x_j) = k(z_i - \bar{z})x_i \cdot \int_{x_j \in (-\infty, +\infty)}(z_j - \bar{z}) \cdot x_j \,d \Pr(x_j)
\]
is proportional to $(z_i - \bar{z})  x_i$ since the integral can be treated as a constant. The second component
\[
    \int_{x_j \in (-\infty, +\infty)} k (z_j - \bar{z}) \cdot (z_i - \bar{z}) \,d\Pr(x_j) = k (z_i - \bar{z}) \cdot \int_{x_j \in (-\infty, +\infty)} (z_j - \bar{z}) \,d\Pr(x_j) = k(z_i - \bar{z}) \Expectation[z_j - \bar{z}]
\]
becomes $0$ due to $ \Expectation[z_j]  = \bar{z}$ . $|S(i)|$ is then proportional to $|x_i|$, thus the data examples around $x=0$ will have smaller absolute values of influence scores.

Then we consider the classifier trained by down-weighting the data examples around $x=0$. We show the case when $|\mu_{1, -1}| < |\mu_{0, +1}|$ in the right figure in Figure \ref{fig:prune_example}. In this case, the down-weighted negative examples from group $z=1$ dominates the down-weighted positive examples from group $z=0$. In consequence, the decision threshold will be perturbed towards right. Coloring the mis-classified examples again, we find out the sum of blue and green area has decreased. The case for $|\mu_{1, -1}| < |\mu_{0, +1}|$ will be symmetric. In conclusion, we demonstrate that removing training examples with smaller absolute influence scores is capable of mitigating the fairness violation.
\end{proof}

\section{Additional Experimental Results}\label{appendix:experiments}
\subsection{Computing Infrastructure}
For all the experiments, we use a GPU cluster with four NVIDIA RTX A6000 GPUs for training and evaluation. We observe that it is efficient to compute the first-order approximated influence scores. It took less than 10 minutes to compute the influence scores for the training examples in CelebA dataset and cost about $0.01$ kg of CO2 equivalent carbon emission.

\subsection{Experimental Details}
\paragraph{Details on Adult dataset}
The UCI Adult \cite{dua2017uci} is a census-based dataset for predicting whether an individual's annual income is greater than 50K, consisting of $14$ features and $48,842$ instances. We select sex, including female and male, as the sensitive attribute. We resample the dataset to balance the class and group membership. The MLP model is a two-layer ReLU network with a hidden size of 64. We train the model 5 times with different random seeds to report the mean and standard deviation of accuracy and fairness metrics. In each trial, the dataset is randomly split into a training and a test set in a ratio of 80 to 20.

\paragraph{Details on CelebA dataset}
The CelebA dataset contains 202,599 face images, where each image is associated with 40 binary human-labeled attributes. We select smiling as binary classification target and gender as the sensitive attribute. We train a ResNet-18 \cite{DBLP:journals/corr/HeZRS15} along with two fully-connected layers for prediction. We follow the original train test splits. We repeat 5 times with different random seeds to report the mean and standard deviation.

\paragraph{Details on Jigsaw dataset}
Jigsaw Comment Toxicity Classification \cite{jigsaw} was initially released as a Kaggle public competition. The task is to build a model that recognizes toxic comments and minimizes the unintended bias with respect to mentions of sensitive identities, including gender and race. 
We select race as the sensitive attribute in our evaluation. Since only a subset provides sensitive identity annotations, we drop the entries with missing race values. We use pre-trained BERT \cite{devlin-etal-2019-bert} to encode each raw comment text into a $768$-dimensional textual representation vector. Then we train a two-layer neural network to perform classification with the encoded textual features. We use the official train set for training and the expanded public test set for testing.

\subsection{Effect of First Order Approximation}
\begin{figure}[htb]
    \centering
    \includegraphics[width=0.6\linewidth]{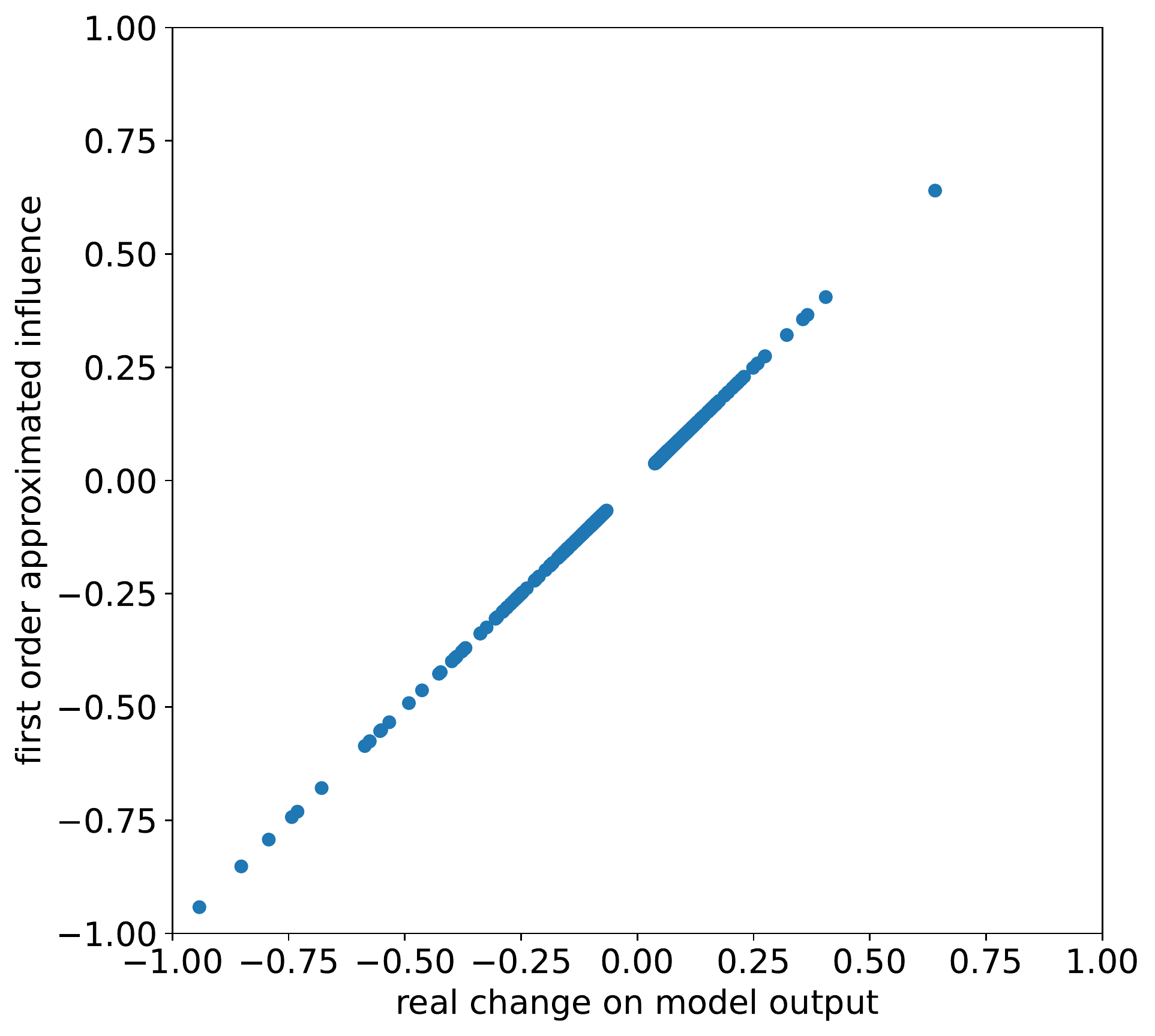}
    \caption{We compare the real change on model output and the pairwise influence with first order approximation.}
    \label{fig:effect-of-first-order-approximation}
\end{figure}

To understand the effect of first order approximation, we train a two-layer neural network with hidden size 64 on the Adult dataset. We randomly pick $1,000$ pairs of training and test points. Upon updating the parameters corresponding to the training point, we calculate the difference of model output on the test point. We directly apply Equation \ref{eq:constrained-influence} to estimate the influence. Figure \ref{fig:effect-of-first-order-approximation} compares how well does the approximated influence align with the real change on model prediction. We find out that the correlation between the two quantities is, without doubt, very closed to 1.

\subsection{Detecting Influential Examples}
\begin{figure}
    \centering
    \subfigure[Accuracy Gap]{
        \includegraphics[height=0.25\linewidth]{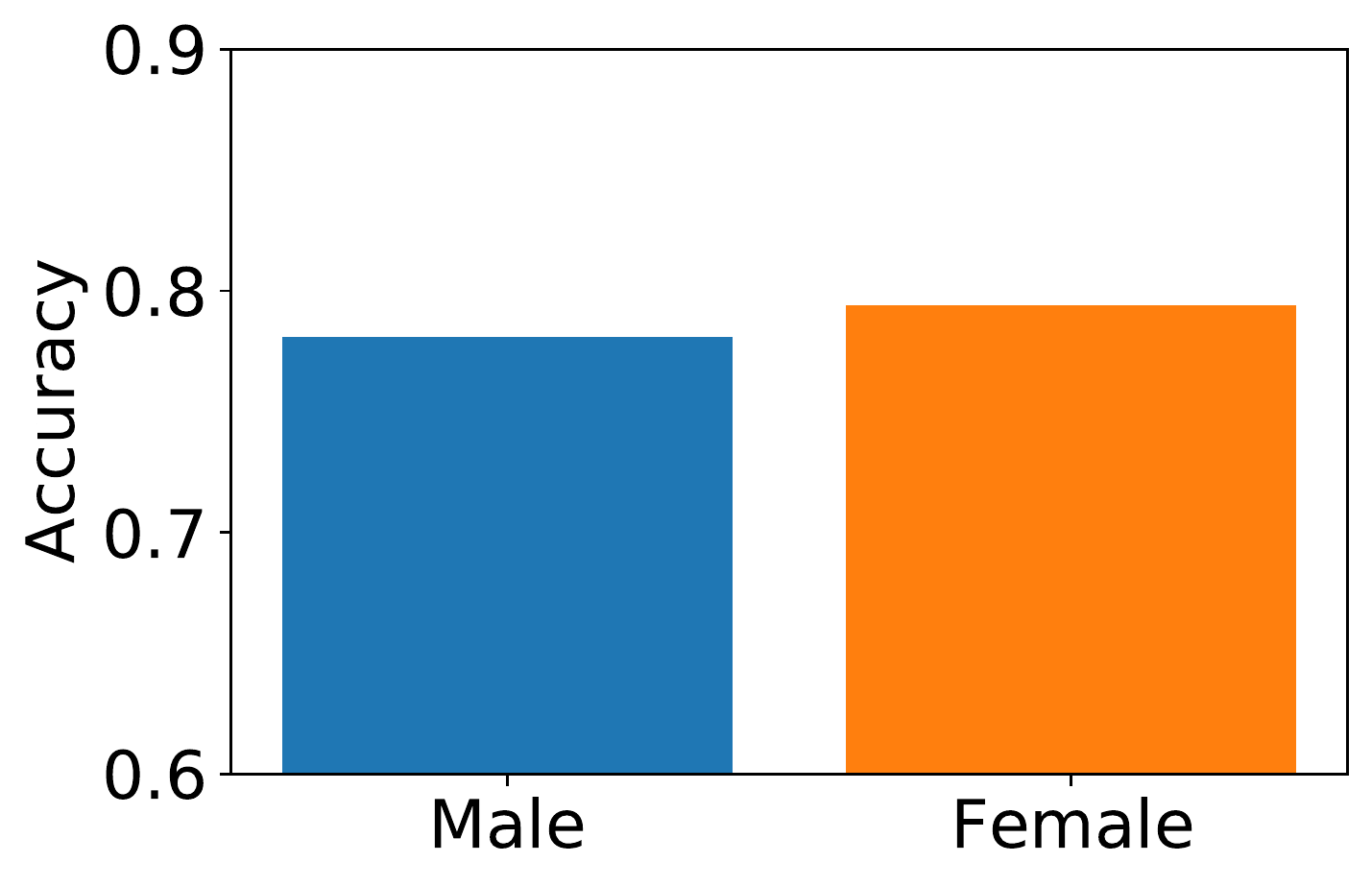}
        \label{fig:acc_cmp}
    }
    \hfill
    \subfigure[Distribution Disparity]{
        \includegraphics[height=0.25\linewidth]{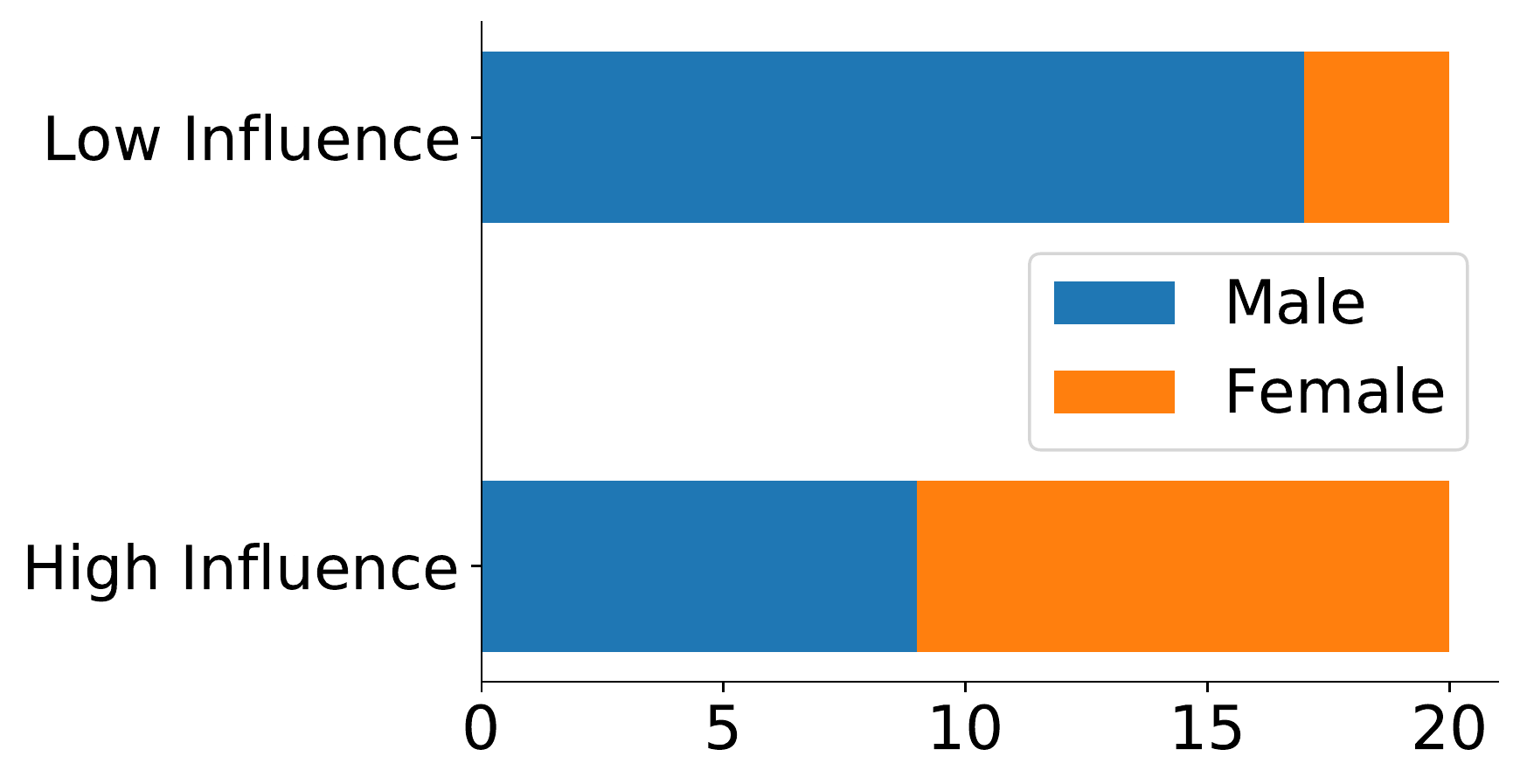}
        \label{fig:head_tail_cmp}
    }
    \caption{We observe that the extracted high influence examples are rather balanced, but the low influence examples are extremely unbalanced for the protected groups.}
\end{figure}

It is intriguing to figure out which examples have the highest and lowest influence scores regarding fairness. On the CelebA dataset, we rank the facial images based on their influence scores by Equation \ref{eq:empirical-influence-score-fairness}. We extract 20 images with highest scores and 20 images with lowest scores, and compare their group distribution in Figure \ref{fig:head_tail_cmp}. We observe that the group distribution is rather balanced for high influence examples, with 9 men and 11 women. However, the group distribution is tremendously skewed towards the male group than the female group, with 17 men and only 3 women. Comparing the accuracy rate for two groups in Figure \ref{fig:acc_cmp}, we conjecture this disparity arose from the lower accuracy for the male group (78.1\%) than the female group (79.4\%).

